\documentclass[conference]{IEEEtran}

\usepackage{amsmath}
\usepackage{amssymb}
\usepackage{graphicx}
\usepackage{booktabs}
\usepackage{multirow}
\usepackage[table]{xcolor}
\usepackage{tikz}
\usepackage{pgfplots}
\usepackage{soul}
\usepackage{makecell}
\usepackage{cite}
\usepackage{url}
\usepackage{balance}
\usetikzlibrary{
  shapes.geometric, arrows.meta, positioning,
  fit, backgrounds, calc,
  matrix, chains
}
\pgfplotsset{compat=1.18}

\tikzset{
  blk/.style={draw, rounded corners=3pt, minimum width=1.7cm,
               minimum height=0.62cm, align=center, font=\scriptsize},
  cnn/.style={blk, fill=blue!9},
  qnn/.style={blk, fill=violet!12},
  cls/.style={blk, fill=green!10},
  fus/.style={blk, fill=orange!11},
  outp/.style={blk, fill=gray!13},
  sat/.style={blk, fill=gray!14, font=\scriptsize},
  gs/.style={blk, fill=blue!9, font=\scriptsize,
             minimum width=2.1cm, minimum height=0.72cm},
  sp/.style={blk, fill=red!10, font=\scriptsize,
             minimum width=1.9cm},
  arr/.style={->, thick, >=Stealth},
  atk/.style={->, dashed, red!65!black, thick, >=Stealth},
  dim/.style={font=\tiny, gray}
}

\begin{document}

\title{\textsc{QuaSAR}: A Quantum-Classical Neural Network for 
SAR Satellite Physical-Layer Authentication}

\author{
    \IEEEauthorblockN{
        Vincenzo Sammartino\IEEEauthorrefmark{1}\IEEEauthorrefmark{2}, 
        Nathanael Denis\IEEEauthorrefmark{2}, and 
        Roberto Di Pietro\IEEEauthorrefmark{2}
    }
    \IEEEauthorblockA{\IEEEauthorrefmark{1}University of Pisa, Pisa, Italy\\
    vincenzo.sammartino@phd.unipi.it}
    \IEEEauthorblockA{\IEEEauthorrefmark{2}King Abdullah University of Science and Technology (KAUST), Thuwal, Saudi Arabia\\
    \{nathanael.denis, roberto.dipietro\}@kaust.edu.sa}
}

\IEEEoverridecommandlockouts
\makeatletter\def\@IEEEpubidpullup{6.5\baselineskip}\makeatother
\IEEEpubid{\parbox{\columnwidth}{
    Network and Distributed System Security (NDSS) Symposium 2027\\
    22--26 March 2027, Seoul, Republic of Korea\\
    ISBN 978-1-970672-09-1\\
    https://dx.doi.org/10.14722/ndss.2027.[23$|$24]xxxx\\
    www.ndss-symposium.org
}
\hspace{\columnsep}\makebox[\columnwidth]{}}

\maketitle

\begin{abstract}
X-band SAR satellites (8--12~GHz) play a critical role in disaster response, environmental monitoring, and military intelligence. Yet, they lack robust physical-layer authentication (PLA), a security layer orthogonal to cryptographic solutions. 
Existing PLA systems, typically based on radio-frequency fingerprinting, are often limited to sub-6~GHz frequencies and rely on classical deep learning. However, this approach underfits the IQ phase nonlinearities that distinguish satellite hardware.
In this paper, we present \textsc{QuaSAR}, 
to the best of our knowledge the first quantum-classical hybrid architecture that fuses a CNN spectrogram encoder with a variational quantum circuit (VQC) to provide PLA 
to X-band SAR signals. Our solution enjoys two distinctive features: (i) it is markedly more data-efficient than classical machine learning, requiring only 10\% of the training data to match the accuracy of classical baselines---data collection being notoriously the most time-consuming phase of PLA; and, (ii) at an equal data budget, it improves classification accuracy over those baselines.
In detail, we test our solution 
under three adversarial scenarios: replay, crafted-IQ injection, and space-borne spoofing. \textsc{QuaSAR} rejects spoofed transmissions in 89.7\%, 94.1\%, and 81.3\% of attempts, respectively, establishing the first quantum-enhanced physical-layer classifier for satellite constellations.
The fully detailed framework and the supporting results, other than being interesting on their own, show a novel research avenue for physical-layer authentication.

\end{abstract}

\IEEEpeerreviewmaketitle

\section{Introduction}

Synthetic aperture radar (SAR) satellites are Earth-observation platforms that transmit active microwave signals, enabling high-resolution surface imaging irrespective of weather or illumination conditions.  The commercial expansion of X-band (8--12~GHz) SAR constellations has accelerated markedly: ICEYE---a Finnish private company owning the world's largest synthetic aperture imaging radar constellation---now operates more than 60~active satellites. Its competitors, Capella Space and Beijing Smart Satellite Space Technology, are similarly scaling deployments. These systems are used for time-critical decisions in disaster management, environmental monitoring, agricultural surveillance, and national security intelligence \cite{Soldi2021, Fontanelli2022, Liu2022, Misra2025}.  Their strategic significance, however, makes them an attractive target for adversaries seeking to forge SAR imagery or disrupt observation services.

Yet, satellite networks do not universally enforce authentication mechanisms~\cite{foruhandeh2020spotr}. Even where cryptographic protections exist, they remain vulnerable to key leakage, legacy hardware incapacity for software updates, and eventual cryptographic compromise over decade-long mission lifetimes~\cite{motallebighomi2022relay}. Physical-layer security (PLS) addresses this vulnerability via an independent mechanism: hardware-induced signal impairments make every radio transmitter physically unique and are extremely hard to reproduce~\cite{sankhe2019no}. While these physical impairments provide a robust authentication anchor, they are subject to drift over extended periods due to the harsh space environment, necessitating periodic model updates via transfer learning to maintain fingerprint accuracy. 

Despite recent progress in RF fingerprinting for sub-6~GHz satellite communications, e.g., IRIDIUM~\cite{Oligeri2023, SmailesIq2025}, the X-band has comparatively received little attention. Commercial SDRs are capped at approximately 6--7~GHz, requiring an intermediate downconversion stage that introduces additional phase noise, mixer nonlinearities, and spurious spectral products~\cite{Syrjala2014}.

Hardware impairments in X-band SAR signals manifest themselves as IQ imbalances and phase micro-perturbations, which are reflected in high-noise radar pulses.  Deep convolutional architectures excel at coarse spectral pattern extraction, but may underfit the nonlinear interactions separating satellites that share hardware generation. Variational quantum circuits (VQCs) offer a theoretically motivated alternative: operating in exponentially large Hilbert spaces with a polynomial parameter count, VQCs can implement high-dimensional nonlinear transformations potentially unavailable to shallow classical layers, amplifying fingerprint separability in directions inaccessible to classical gradient
descent~\cite{biamonte2017quantum}.



\textbf{Contributions.} This paper provides a novel PLA mechanism for satellite authentication that blends quantum and classical ML techniques. 
In particular, we provide the following contributions:

\begin{itemize}
  \item We design and deploy an X-band SAR satellite RF
    fingerprinting testbed, based on a programmable downconversion mixer and a USRP X310~SDR. To validate our framework, we collect 3.76~TB of raw IQ data from 37~operational ICEYE satellites over 28~days using two independent SDR units---the latter to assess cross-receiver transferability.

  \item We propose \textsc{QuaSAR}, a quantum-classical hybrid
    architecture combining a deep CNN spectrogram encoder, a
    four-layer VQC operating over an 8-qubit register, and a
    classical skip connection fused via late concatenation.
    The VQC is driven by an IQ-native encoding that maps the
    amplitude and the phase of each projected feature onto the polar
    and azimuthal angles of a qubit, exploiting the geometric
    isomorphism between a complex sample and a single-qubit pure
    state.

  \item We demonstrate 97.3\% validation accuracy and macro-F1\,=\,0.973
    in binary satellite authentication using only 10\% of the collected corpus---matching classical baseline accuracy at a fraction of the enrollment cost---and exceeding the classical-only baseline by 7.5~percentage points. Furthermore, through a dedicated explainability analysis based on gradient saliency maps and latent-space clustering, we establish the physical grounding of \textsc{QuaSAR}'s decisions, confirming the ability of the quantum branch to amplify minute hardware fingerprints and localizing the decision to sub-millisecond windows in the raw IQ domain.

  \item We define and test  three attack scenarios:
    replay attacks; crafted-IQ injection; and, 
    spoofing---achieving detection rates of 
    89.7\%, 94.1\%, and 81.3\%, respectively.
\end{itemize}

The remainder of this paper is organized as follows.
Section~\ref{sec:bg} provides background.
Section~\ref{sec:threat} defines the threat model.
Section~\ref{sec:data} describes data collection and processing.
Section~\ref{sec:arch} presents \textsc{QuaSAR}.
Section~\ref{sec:eval} reports experimental results.
Section~\ref{sec:explain} provides explainability analysis.
Section~\ref{sec:related} surveys related work.
Section~\ref{sec:disc} discusses limitations and future work. Conclusions are reported in Section~\ref{sec:concl}.

\section{Background}
\label{sec:bg}
This section introduces the principles underlying our proposed approach, specifically SAR satellite imaging, software-defined radio for RF fingerprinting, time-frequency signal representations, and variational quantum circuits. We also connect IQ samples to qubits to further justify the use of quantum machine learning for PLA.

\subsection{SAR Imaging Satellites}

Synthetic Aperture Radar (SAR) is an active microwave remote sensing technology. A SAR satellite illuminates the ground with radio pulses and reconstructs a high-resolution image from the reflected echoes. Because the system relies on its own illumination and operates at microwave frequencies that penetrate clouds, SAR acquires imagery day or night, independently of weather. Passive optical satellites, by contrast, offer no such capability, which makes SAR the instrument of choice for disaster response, maritime surveillance, and military intelligence \cite{Soldi2021, Lv2023, Misra2025}.

\textbf{The imaging pulse.} To achieve fine range resolution without requiring impractically short pulses, SAR systems transmit a linearly frequency-modulated (LFM) chirp: a waveform whose
instantaneous frequency sweeps linearly across the instrument
bandwidth over the pulse duration \cite{Moreira2013}. For a chirp of bandwidth $B$ and
duration $T_p$, the baseband transmitted signal is
\begin{equation}
  s(t) = \mathrm{rect}\!\left(\tfrac{t}{T_p}\right)
         \exp\!\bigl(j\pi K t^2\bigr),
  \qquad K = B/T_p,
  \label{eq:chirp}
\end{equation}
where $K$ is the chirp rate \cite{Curlander1991, Moreira2013}. Pulse compression at the receiver trades
the long transmitted pulse for a short effective pulse of width
$\approx\!1/B$, so ground range resolution scales inversely with the
transmitted bandwidth. 

\textbf{SAR satellite imaging bands.}
SAR constellations operate across several microwave bands---L, C, and X being the most common \cite{Reigber2002, Bonano2013, Cyprien2024}. The X-band (8--12~GHz) offers the most favorable trade-off between resolution and antenna size for small commercial platforms \cite{Bonano2013}: its short wavelength yields sub-meter ground resolution with antennas compatible with comparatively lightweight satellite buses, i.e., even below 100 kg, enabling the dense LEO constellations now deployed by ICEYE, Capella Space, and Umbra. Lower bands (L, C) require proportionally larger antennas for equivalent resolution and are typically reserved for flagship government missions \cite{Golkar2021}.

\textbf{The ICEYE constellation.}
Our testbed targets satellites from the ICEYE constellation. ICEYE is a Finnish commercial operator,
founded in 2014 as a spin-off of Aalto University, whose mission was to miniaturize X-band SAR payloads to platforms below 100~kg. Each ICEYE satellite carries an active phased-array antenna operating at a nominal carrier frequency of 9.65~GHz and supports several acquisition modes --- Strip, Spot,
Scan, and Dwell ---, with published ground resolutions ranging from
approximately 0.25~m to 15~m depending on mode \cite{eoportal2026iceye}. ICEYE now operates the largest commercial SAR constellation in orbit; the 37~satellites used in this study correspond to the subset operational during our 28-day collection campaign, while the constellation has since grown to over 60~satellites. Within a single hardware generation, all satellites share a common bus and payload design, which makes intra-constellation discrimination particularly challenging for any fingerprinting method.

\subsection{Software-Defined Radio and RF Fingerprinting}
SDRs implement signal processing on reconfigurable hardware, enabling passive wideband capture without prior knowledge of the target's transmission scheme.  Their operational frequency ceiling (6--7~GHz for commodity units) precludes direct capture of X-band emissions.  Radio fingerprinting exploits
hardware impairments to identify the originating transmitter~\cite{sankhe2019no}.  Even mass-produced transceivers from the same batch are distinguishable due to minute differences in hardware, which manifest as IQ gain imbalance, DC offset, phase noise, and nonlinear harmonic distortion in received IQ samples~\cite{Oligeri2023}.

\textbf{Short-Time Fourier Transform spectrograms.}
Raw IQ samples are converted to time-frequency representations
via the STFT \cite{Nawab1983}.  For a complex baseband signal $x[n]$:
\begin{equation}
  X(m,k) = \sum_{n=0}^{N-1} x[n+mH]\,w[n]\,e^{-j2\pi kn/N}
  \label{eq:stft}
\end{equation}
where $w[n]$ is a Hanning analysis window of length~$N$, $H$ is
the hop size, and $m,k$ index time frames and frequency bins,
respectively.  The log-magnitude spectrogram
$S(m,k)=10\log_{10}|X(m,k)|^2$ is treated as a grayscale 2D
image, jointly encoding spectral content and temporal evolution.

\subsection{Variational Quantum Circuits}

A VQC $U(\boldsymbol{\theta},\mathbf{x})$ encodes classical data into qubit states, applies a parameterized gate sequence, and returns expectation values of Pauli observables as classical outputs. An angle-embedding layer initializes each qubit via:
\begin{equation}
  R_Y(\phi_i)|0\rangle,\quad \phi_i = \pi\cdot\sigma(x_i),\quad i=1,\ldots,d
  \label{eq:embed}
\end{equation}
where $\sigma(\cdot)$ is the sigmoid function, constraining $\phi_i\in(0,\pi)$. 
The embedding loads each classical feature onto an independent qubit, producing a separable product state $\bigotimes_{i=1}^{d}|\phi_i\rangle$ that carries no inter-qubit correlation. To learn joint functions of the embedded features, this state is processed by $L$ strongly entangling layers (SEL)~\cite{schuld2020circuit}.  Each SEL applies parameterized $R_Y/R_Z$ rotations on every qubit followed by a  CNOT ring, generating the multi-qubit entanglement needed to evaluate non-separable functions over the $2^d$-dimensional Hilbert space. The circuit output is the vector of single-qubit Pauli-$Z$ expectation values:
\begin{equation}
  \mathbf{v}_Q = \bigl[\langle Z_1\rangle,\langle Z_2\rangle,\ldots,\langle Z_d\rangle\bigr] 
  \in [-1,1]^d,
  \label{eq:vqc}
\end{equation}
which is differentiable end-to-end via the parameter-shift rule~\cite{mitarai2018quantum}, 
making the VQC compatible with standard backpropagation.

\subsection{Connecting IQ Samples to Qubits}
\label{ss_iq_to_qubit}

\begin{figure}[t]
    \centering
    \includegraphics[width=\linewidth]{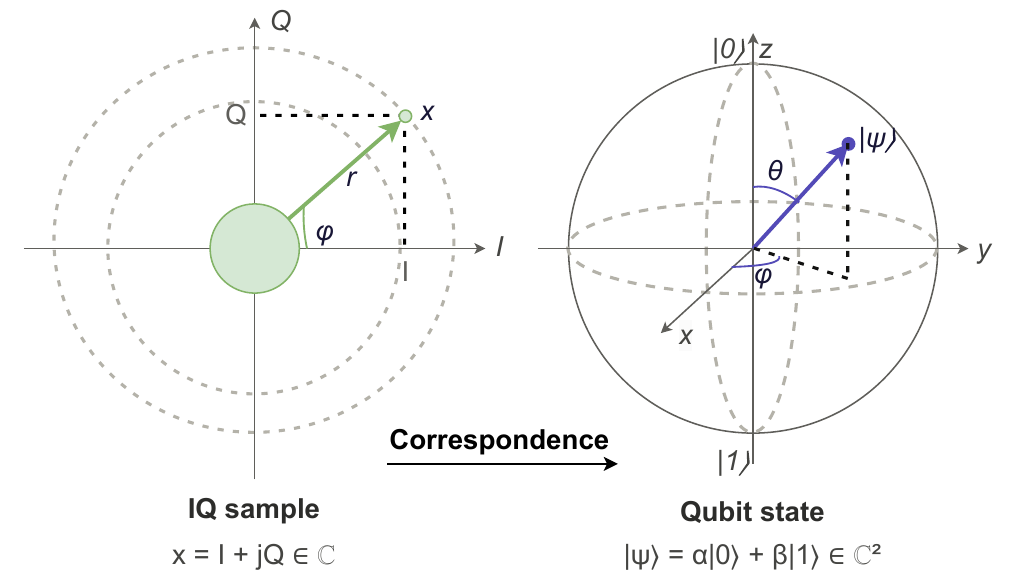}
    \caption{\textbf{IQ samples and qubit states.}
    Left: a complex IQ sample $x = I + jQ$, represented as a vector in the complex plane with magnitude $r$ and phase $\varphi$. Right: a single-qubit pure state $|\psi\rangle = \cos(\theta/2)|0\rangle + e^{i\varphi}\sin(\theta/2)|1\rangle$ on the Bloch sphere, parameterized by the polar angle $\theta$ and the azimuthal angle $\varphi$. Both objects are fully specified by two real parameters: the IQ magnitude $r$ maps to the polar angle $\theta = 2\arcsin(r)$, while the phase $\varphi$ is the same physical quantity in both representations.}
    \label{fig:iq_qubit}
\end{figure}

A central motivation for using Quantum Machine Learning (QML) in the context of radio-frequency fingerprinting is that the native state space of a qubit is geometrically isomorphic to the native state space of an IQ sample. This correspondence, illustrated in Figure~\ref{fig:iq_qubit}, permits an encoding under which the entire information content of a complex sample, i.e., both amplitude and phase, is preserved in the quantum state. This is often not true for real-valued encodings, e.g., of spectrogram magnitudes, which discard phase by construction.

\textbf{The geometric correspondence.}
A complex IQ sample $x = I + jQ \in \mathbb{C}$ is uniquely determined by two real parameters: its magnitude $r = |x|$ and its phase $\varphi = \arg(x) \in [0, 2\pi)$. A single-qubit pure state
\begin{equation}
  |\psi\rangle = \cos(\theta/2)\,|0\rangle
               + e^{i\varphi}\sin(\theta/2)\,|1\rangle
  \label{eq:qubit_state}
\end{equation}
is likewise determined by two real parameters: the polar angle
$\theta \in [0, \pi]$ and the azimuthal angle $\varphi \in [0, 2\pi)$ on the Bloch sphere. The azimuthal angle $\varphi$ is the same physical quantity in both representations, i.e., the phase of
the complex object. Only the amplitude--polar mapping needs to be specified to complete the correspondence.

\textbf{IQ-native encoding.}
Given a normalized IQ sample $x_n = I_n + jQ_n$ with
$r_n = |x_n| \in [0,1]$ and $\varphi_n = \arg(x_n)$, we encode it
into qubit $n$ via the two-rotation sequence
\begin{equation}
  |\psi_n\rangle \;=\; R_Z(\varphi_n)\,R_Y(\theta_n)\,|0\rangle,
  \qquad \theta_n = 2\arcsin(r_n),
  \label{eq:iq_encoding}
\end{equation}
where $R_Y(\theta) = \exp(-i\theta Y/2)$ and
$R_Z(\varphi) = \exp(-i\varphi Z/2)$ are the standard Pauli
rotations \cite{Nielsen2010, schuld2021ml}. The $R_Y$ rotation fixes the latitude of the state on
the Bloch sphere (loading the amplitude), while the subsequent
$R_Z$ rotation fixes its longitude (loading the phase). The $R_Z$ rotation introduces an overall factor $e^{-i\varphi/2}$ which, as a global phase, leaves all measurable quantities and subsequent gate operations invariant. Excluding this global phase, the resulting state is exactly
Eq.~\eqref{eq:qubit_state} with $\theta = \theta_n$ and
$\varphi = \varphi_n$, so the encoding is lossless: the original sample can be recovered as $x_n = \sin(\theta_n/2)\,e^{i\varphi_n}$ from the Bloch-sphere coordinates of $|\psi_n\rangle$. 

A block of $d$ consecutive IQ samples is encoded into a $d$-qubit product state $|\psi\rangle = \bigotimes_{n=1}^{d} |\psi_n\rangle$ by applying Eq.~\eqref{eq:iq_encoding} independently on each qubit. The entangling layers that follow (cf. Section~\ref{sec:arch}) subsequently couple these qubits, allowing the VQC to learn joint functions of the full IQ block.

\textbf{Contrast with angle embedding of real features.}
The conventional angle-embedding approach of
Eq.~\eqref{eq:embed} loads a real scalar $x_i \in \mathbb{R}$ into
qubit $i$ via a single $R_Y$ rotation, leaving the azimuthal
degree of freedom unused. Applied to a complex IQ stream, this
scheme requires either: (i) discarding the phase; or, (ii) spending
two qubits per sample (one for $I$, one for $Q$), which doubles
circuit depth and breaks the geometric link between the encoded
state and the original complex number. The IQ-native encoding of
Eq.~\eqref{eq:iq_encoding} avoids both penalties: one sample maps
to one qubit with no information loss, while  the phase
$\varphi$---a feature that may carry hardware-induced
impairments such as oscillator drift and mixer
nonlinearity \cite{Agadakos2020, Jun2022}--- is represented on the qubit as the same angle it occupies in the complex plane. Section~\ref{ss:encoding_ablation} quantifies what this
distinction is worth in practice: replacing angle embedding with the
IQ-native encoding, with every other component of the architecture
held fixed, raises test accuracy by 2.2~percentage points and nearly
halves the number of epochs to convergence.

\section{Threat Model}
\label{sec:threat}

\begin{figure}[t]
\centering
\includegraphics[width=\columnwidth]{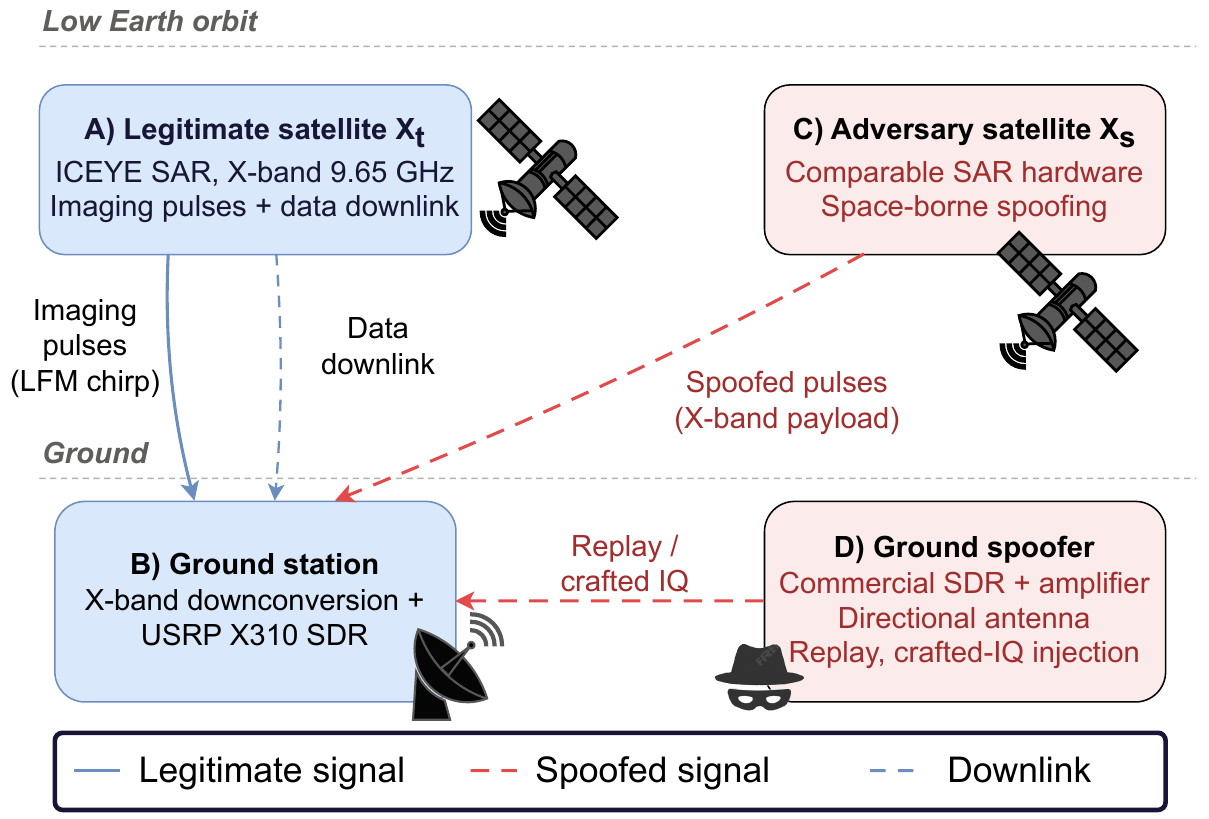}
\caption{%
  \textbf{Threat model.}
  (A) A legitimate SAR satellite $X_t$ transmits two
  physically distinct X-band signals: i) (frequent) high-power chirp pulses
  for Earth imaging; and, ii) (uncommon) phase-modulated data downlinks.
  (B) A ground station captures imaging pulses for authentication purposes.
  Two adversaries are considered: (C) a
  space-borne adversary operating a SAR satellite $X_s$ of
  comparable hardware; and, (D) a ground spoofer operating
 a commercial high-end SDR.}
\label{fig:threat}
\end{figure}

We define adversarial assumptions, objectives, and attack scenarios that scope the design of \textsc{QuaSAR}. Figure~\ref{fig:threat} illustrates the system and threat model.

\subsection{System model}
A ground station equipped with an X-band downconversion chain and SDR passively captures radar illuminating pulses transmitted by a legitimate ICEYE SAR satellite during orbital passes. A trained \textsc{QuaSAR} classifier authenticates illumination signals against the enrolled satellite's physical-layer fingerprint in near real time. Enrollment is performed offline on verified satellite passes.

\textbf{Adversarial objective.}
The adversary seeks to compromise physical-layer authentication, either by successfully impersonating a legitimate satellite to inject fabricated IQ data that generates false SAR imagery or
by blocking correct authentication to deny observation services. In high-stakes contexts, such as military surveillance or critical-infrastructure monitoring, both outcomes carry severe operational consequences.

\textbf{Adversarial capabilities.}
We consider a tiered adversary model whose resources scale with the attack scenario. At the lower tier, a terrestrial attacker possesses a commercial off-the-shelf SDR, a power amplifier, and a directional antenna, enabling signal capture, replay, and synthesis in proximity to the target ground station. This
equipment is commercially available for \$500--\$10{,}000~USD and is sufficient for ground-based replay and crafted-IQ attacks. At the upper tier, we consider a strategic adversary with the means to operate an X-band SAR satellite of their own. While orders of magnitude more expensive, this capability is no longer the exclusive domain of major space powers: commercial SAR satellites can be procured or leased from established vendors. Rideshare launches, where several customers share one rocket, have driven costs into the low millions of USD range for a mission. As a consequence, several nation-states as well as non-state actors can now field X-band imaging assets. There are several incentives for a strategic adversary. Fabricating or suppressing SAR imagery can conceal military deployments, manipulate insurance and commodity markets that rely on SAR-derived intelligence, undermine adversarial disaster-response or treaty-verification efforts, or discredit a competing constellation operator by injecting falsified observations attributed to their satellites. In all tiers, we do not assume access to the legitimate ICEYE satellite hardware, knowledge of \textsc{QuaSAR}'s parameters, or compromise of the ground station.

\subsection{Attack scenarios}

We consider three scenarios, ordered by increasing adversarial capability. They are referred to as Scenarios~(A), (B), and~(C) throughout the paper.

\textbf{Scenario (A): replay.}
The adversary, equipped with a commercial SDR, a power amplifier, and a directional antenna, records IQ bursts during a legitimate pass of $X_t$ and re-emits them from the ground toward the victim ground station during a subsequent window. This is the weakest of the three models in terms of required expertise and capital outlay (\$500--\$10{,}000~USD), but also the most operationally feasible. The defense relies on the fact that the spoofer's transmit chain superimposes its own hardware impairments on top of $X_t$'s original fingerprint, producing a measurable discrepancy in the IQ domain. We instantiate this scenario by replaying previously captured ICEYE bursts through a second USRP~X310 transmitting at the IF stage of our acquisition chain.

\textbf{Scenario (B): crafted-IQ injection.}
Rather than replaying a recorded waveform, the adversary synthesizes an LFM chirp that matches the nominal ICEYE waveform parameters---carrier, bandwidth, chirp rate, and pulse repetition interval---and transmits it toward the ground station. The synthetic burst is spectrally indistinguishable from a legitimate illumination at the macro level, but carries no hardware fingerprint of $X_t$: only the impairments of the adversary's own transmit chain. This scenario isolates the classifier's reliance on micro-scale impairments rather than on waveform structure.

\textbf{Scenario (C): space-borne spoofing.}
A second SAR satellite $X_s$ operates in the same band and shares comparable hardware with the legitimate satellite $X_t$. $X_s$ transmits illuminating pulses to the ground station, aiming to spoof $X_t$. In a realistic deployment, $X_s$ would belong to a foreign or competing constellation; however, such cross-constellation captures are not available in our dataset, and an adversary controlling a satellite from a different hardware family would be easier to discriminate due to larger impairment gaps. To stress-test \textsc{QuaSAR} under the challenging space-borne conditions, we therefore emulate this scenario via an \emph{open-set} protocol: a subset of the 37~ICEYE satellites is held out from the training and validation
splits and presented to the classifier exclusively at test time in the role of $X_s$. Because ICEYE units share similar phased-array architecture and hardware, this is a plausible space-borne attacker. A successful defense requires the model to reject $X_s$ despite minimal macro-level spectral divergence
from~$X_t$. This is the strongest of the three models, and the one against which \textsc{QuaSAR} is hardest pressed.

\section{Data Collection and Processing}
\label{sec:data}

This section describes the hardware testbed used to capture X-band SAR signals, the rationale for targeting imaging pulses over data downlinks, the resulting dataset, and the preprocessing pipeline that converts raw IQ recordings into spectrograms for further processing.

\subsection{Acquisition Testbed}

\begin{figure}
    \centering
    \includegraphics[width=\linewidth]{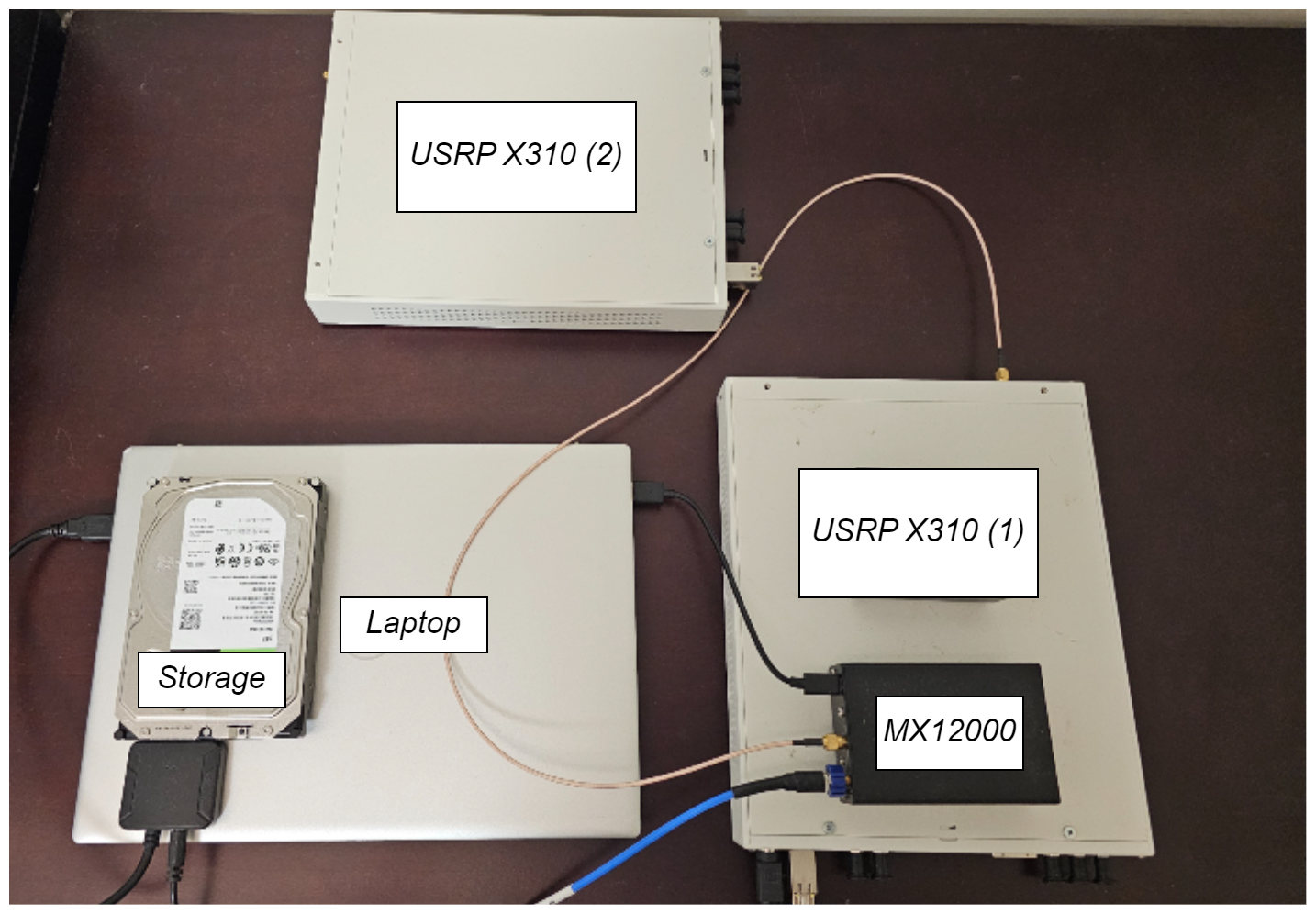}
    \caption{Setup to collect satellite signals in the X-band. Two USRPs are used, along with an MX12000 downconverter and 8TB Seagate BarraCuda storage. }
    \label{fig:testbed}
\end{figure}

Capturing ICEYE SAR illuminating signals at 9.65~GHz lies beyond the 6--7~GHz ceiling of commodity SDRs.  We designed a custom downconversion chain composed of three elements. The Aaronia Hyperlog
Pro 70140 is a passive directional antenna with a wide frequency range spanning from 700 MHz to 14 GHz, a 13 dBi gain, and is designed for field use. It feeds a DSI MX12000 programmable microwave mixer with a local oscillator set to 7.2~GHz, downconverting the received signal to an intermediate frequency (IF) of
2.45~GHz. The MX12000 eliminates the need for an external high-frequency signal source. The MX12000 output is captured by an Ettus USRP~X310 equipped with UBX-40 daughterboards (10~MHz--6~GHz, 40~MHz instantaneous bandwidth) at 10~Msps. IQ files are stored on an 8~TB HDD and processed offline using GNU Radio.

The 40~MHz capture bandwidth is narrower than ICEYE's full imaging bandwidth --- up to 299~MHz for standard modes. Full-bandwidth capture is required for SAR image formation, but not for fingerprinting. Indeed, hardware impairments are detectable within any consistent spectral slice, so fingerprinting remains viable within the SDR's instantaneous bandwidth.

\subsection{Rationale for Targeting Imaging Pulses}
Our testbed exclusively captures SAR imaging pulses rather than
the satellite's X-band data downlink. Three properties make imaging pulses the natural choice for physical-layer authentication.

\textbf{Passive availability.}
Imaging pulses are high-power illuminations broadcast toward Earth during every orbital pass, and can be collected by any ground receiver with direct line-of-sight visibility to the satellite. The data downlink, by contrast, is a directed high-gain beam steered toward a specific licensed ground station, making passive interception impractical without prior coordination, and in particular, proximity to the intended receiver.

\textbf{Signal richness.}
Imaging pulses are wideband LFM chirps, up to 299~MHz for ICEYE standard modes, which are transmitted at high power. They provide a large spectral canvas over which hardware impairments can manifest. The data downlink employs narrowband phase-modulated communication waveforms, e.g., QPSK or 8PSK, whose reduced
bandwidth and distinct modulation scheme produce a fundamentally different impairment signature. As a result, a fingerprint trained on one cannot substitute for the other.

\textbf{Collection frequency.}
Imaging pulses are emitted continuously throughout the illumination phase of every pass, whereas data downlink sessions are comparatively infrequent and tied to the availability of licensed downlink stations. This asymmetry substantially facilitates dataset collection, both for research purposes---where independent ground receivers can accumulate large corpora across many passes---and for operational deployment by a constellation operator equipped with its own ground infrastructure.

Note that the hardware fingerprint extracted from SAR imaging pulses is not interchangeable with one derived from the satellite's X-band data downlink, as the two signals differ fundamentally in waveform structure.

\subsection{Dataset}

A 28-day collection campaign produced signals from all 37~operational ICEYE satellites, accumulating 3.76~TB of raw IQ data. To assess cross-receiver transferability---a known source of fingerprinting bias~\cite{irfan2024reliability}---two physically separate USRP~X310 units were deployed: the first
unit contributed 2.48~TB and the second~1.28~TB.

\subsection{Pre-processing}
\label{ss_pre_processing}
Noise-only intervals are mitigated via an adaptive magnitude threshold:
\begin{equation}
  \tau = \alpha\cdot(\mu + 2\sigma)
  \label{eq:thresh}
\end{equation}
where $\mu$ and $\sigma$ are the mean and standard deviation of instantaneous sample magnitudes and $\alpha\in[0,1]$ is a tunable strength coefficient.  Samples exceeding $\tau$ are retained while the remainder are discarded, reducing the noise floor during downconverted X-band reception.

\subsection{Spectrogram Generation}

Retained IQ bursts are segmented into chunks of 100{,}000~samples. Each chunk is transformed into a grayscale spectrogram using Eq.~\eqref{eq:stft} with Hanning window $N=256$, hop $H=128$
(50\% overlap), and FFT length~256.  The power spectrogram is log-scaled in dB, clipped between the 1st and 99th percentiles for dynamic-range compression, and resized to $224\times224$ pixels to match the CNN encoder's input dimensions.

\subsection{Class Balancing}
We collect IQ data from 37 ICEYE satellites and train one binary classifier per satellite under a one-vs-rest (OvR) protocol. For each classifier, samples from the enrolled satellite are the positive class and samples from the remaining 36 satellites form the negative class. This creates a natural $36{:}1$ imbalance. To correct it, we undersample the negative class at the pass level, drawing negative passes until their spectrogram count matches the positive class, and close any residual gap with ceiling oversampling. The final balanced set contains 2{,}021 spectrograms per classifier, evenly split between the two classes. This procedure is applied independently for each of the 37 enrolled satellites.

\begin{figure}[t]
\centering
\includegraphics[width=\columnwidth]{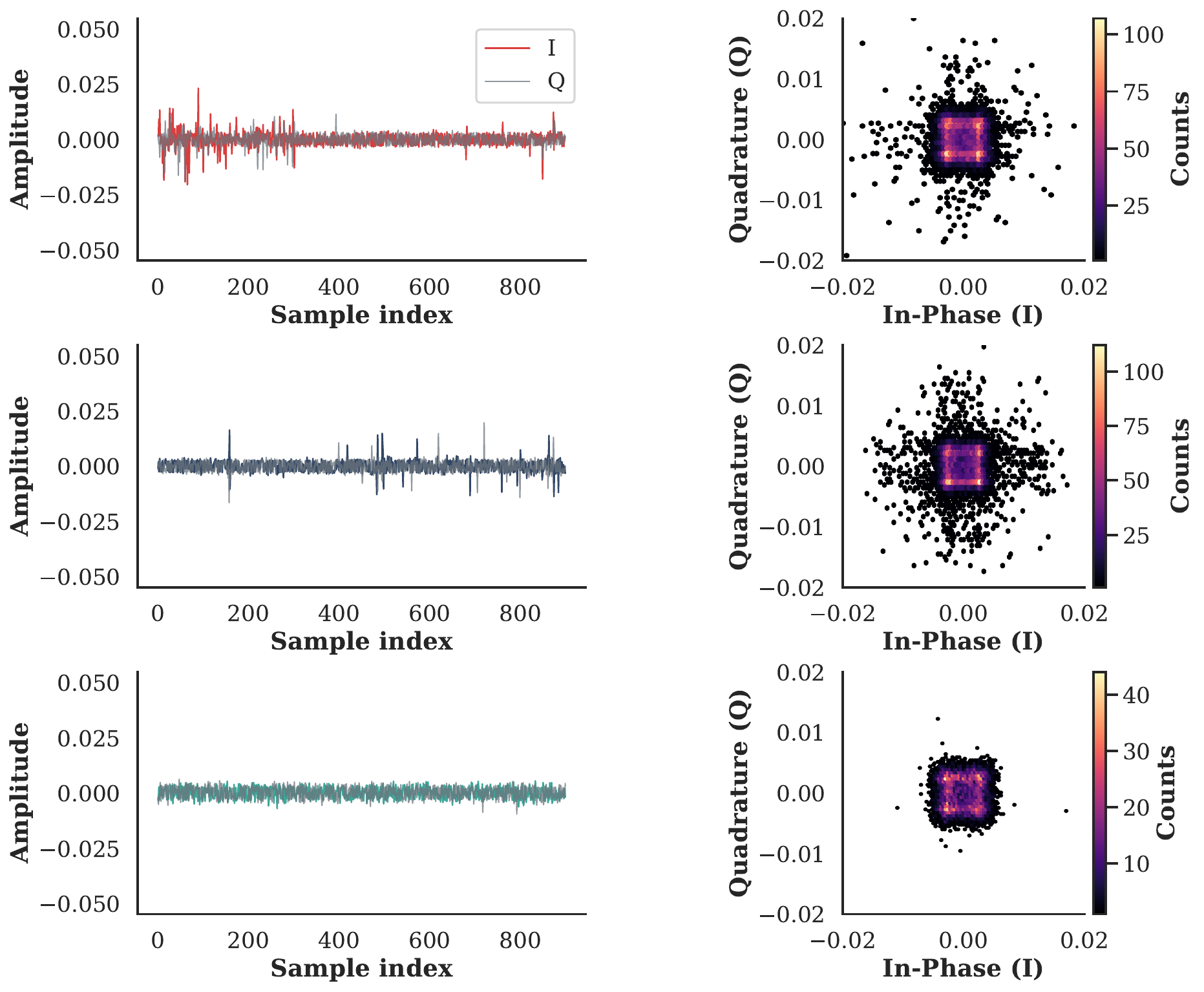}
\caption{Time-domain IQ waveforms (left) and constellation
  density maps (right) for three ICEYE satellites:
  X11 (top), X13 (middle), and X14 (bottom).}
\label{fig:iq_morphology}
\end{figure}

Each satellite is captured on its own orbital passes, in separate
acquisitions; no capture contains more than one emitter. The labels
\emph{target} and \emph{non-target} are therefore not properties of a
collection but roles assigned per classifier: under the one-vs-rest
protocol of Section~\ref{sec:data}, each of the 37~satellites is enrolled
as the target in turn, and the remaining 36 take the non-target role for
that classifier. Figure~\ref{fig:iq_morphology} shows three of these
independently acquired emitters, with X11 drawn in the target role purely
for illustration.

The three rows illustrate the morphological diversity the testbed
resolves. X11 (top row) exhibits pronounced burst-onset transients in
the time-domain trace, with I and Q components reaching peak amplitudes
of $\pm$0.05, followed by rapid signal decay. Its constellation density
map is broad and asymmetric, reflecting the hardware phase noise and IQ
imbalance introduced by that satellite's transmit chain. X13 (middle
row) displays sparse impulsive artifacts superimposed on a
lower-amplitude stationary floor, with a wider constellation
spread indicating a noise process with higher variance.  X14 (bottom
row) presents the most stationary
waveform, with uniformly bounded amplitude and a compact,
near-circular constellation whose reduced count density
($\leq$40 samples at peak) reflects its intermittent in-band
activity.  These inter-emitter differences---detectable at the
level of raw IQ morphology prior to any spectral
transformation---confirm that the downconversion chain
preserves sufficient hardware-induced signal structure to
support fingerprint extraction, and motivate the class-balancing
procedure described above.

\subsection{Training / Validation / Test Split}

The balanced dataset is partitioned into $D_\mathrm{train}$ (60\%),
$D_\mathrm{val}$ (20\%), and $D_\mathrm{test}$ (20\%) at the granularity of orbital passes rather than of individual spectrograms: we first group all bursts by pass identifier (date $\times$ satellite $\times$ acquisition window), then assign whole passes to the three subsets by random shuffling. Spectrograms from the same pass therefore never straddle two subsets. This prevents temporally adjacent bursts---which share channel state, atmospheric conditions, and front-end thermal regime---from leaking across splits.

\section{\textsc{QuaSAR}: Architecture}
\label{sec:arch}

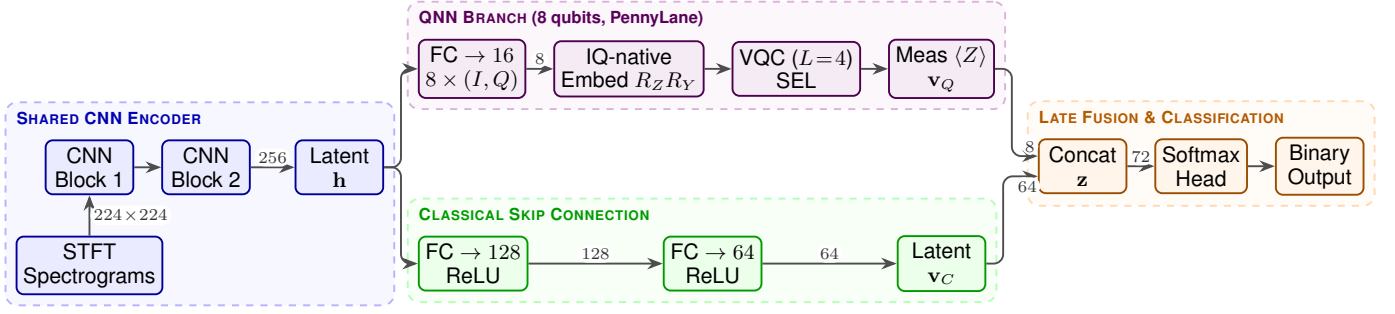
\begin{figure*}[t]
\centering
\resizebox{\linewidth}{!}{%
\begin{tikzpicture}[
  node distance=0.6cm and 0.4cm,
  >=Stealth,
  box/.style={
    rectangle, draw, thick, align=center,
    inner sep=0.1cm, minimum width=1.3cm, minimum height=0.8cm,
    font=\sffamily\small, rounded corners=3pt
  },
  cnn/.style={box, fill=blue!8, draw=blue!60!black},
  qnn/.style={box, fill=violet!8, draw=violet!60!black},
  cls/.style={box, fill=green!8, draw=green!60!black},
  fus/.style={box, fill=orange!8, draw=orange!60!black},
  arr/.style={->, thick, draw=black!70, rounded corners=4pt},
  lbl/.style={
    font=\sffamily\scriptsize, fill=white, inner sep=1.5pt, text=black!80
  },
  bg/.style={
    rounded corners=5pt, dashed, line width=0.8pt, inner sep=0.15cm
  },
  bg_enc/.style={bg, fill=blue!3, draw=blue!30},
  bg_qnn/.style={bg, fill=violet!3, draw=violet!30},
  bg_cls/.style={bg, fill=green!3, draw=green!30},
  bg_fus/.style={bg, fill=orange!3, draw=orange!30},
  bgtitle/.style={
    font=\sffamily\bfseries\scriptsize, text=#1
  }
]

  \node[cnn] (inp) {STFT\\Spectrograms};
  \node[cnn, above=0.6cm of inp] (c1) {CNN\\Block 1};
  \node[cnn, right=0.4cm of c1] (c2) {CNN\\Block 2};
  \node[cnn, right=0.65cm of c2] (lat) {Latent\\$\mathbf{h}$};

  \draw[arr] (inp) -- node[lbl, right] {$224\!\times\!224$} (c1);
  \draw[arr] (c1)  -- (c2);
  \draw[arr] (c2)  -- node[lbl, above] {$256$} (lat);

  \node[qnn, above right=0.6cm and 0.5cm of lat] (proj) {FC $\to 16$\\$8\times(I,Q)$};
  \node[qnn, right=0.4cm of proj] (emb) {IQ-native\\Embed $R_ZR_Y$};
  \node[qnn, right=0.4cm of emb] (vqc) {VQC ($L\!=\!4$)\\SEL};
  \node[qnn, right=0.4cm of vqc] (meas) {Meas $\langle Z \rangle$\\$\mathbf{v}_Q$};

  \draw[arr] (proj) -- node[lbl, above] {$8$} (emb);
  \draw[arr] (emb)  -- (vqc);
  \draw[arr] (vqc)  -- (meas);

  \node[cls, below right=0.6cm and 0.5cm of lat] (fc1) {FC $\to 128$\\ReLU};
  \node[cls] at (meas |- fc1) (vc) {Latent\\$\mathbf{v}_C$};
  \path (fc1) -- (vc) node[midway, cls] (fc2) {FC $\to 64$\\ReLU};

  \draw[arr] (fc1) -- node[lbl, above] {$128$} (fc2);
  \draw[arr] (fc2) -- node[lbl, above] {$64$} (vc);

  \draw[arr] (lat.east) -- ++(0.25,0) |- (proj.west);
  \draw[arr] (lat.east) -- ++(0.25,0) |- (fc1.west);

  \coordinate (mid_right) at ($(meas.east)!0.5!(vc.east)$);
  \node[fus, right=0.7cm of mid_right] (cat) {Concat\\$\mathbf{z}$};
  \node[fus, right=0.4cm of cat] (head) {Softmax\\Head};
  \node[fus, right=0.4cm of head] (out) {Binary\\Output};

  \draw[arr] (cat)  -- node[lbl, above] {$72$} (head);
  \draw[arr] (head) -- (out);

  \draw[arr] (meas.east) -- ++(0.25,0) |- node[lbl, above, pos=0.85] {$8$} ([yshift=0.15cm]cat.west);
  \draw[arr] (vc.east)   -- ++(0.25,0) |- node[lbl, below, pos=0.85] {$64$} ([yshift=-0.15cm]cat.west);

  \coordinate (pad_enc) at ($(lat.north) + (0, 0.4)$);
  \coordinate (pad_qnn) at ($(proj.north) + (0, 0.4)$);
  \coordinate (pad_cls) at ($(fc1.north) + (0, 0.4)$);
  \coordinate (pad_fus) at ($(cat.north) + (0, 0.4)$);

  \begin{scope}[on background layer]
    \node[bg_enc, fit=(inp) (lat) (pad_enc)] (bg1) {};
    \node[bg_qnn, fit=(proj) (meas) (pad_qnn)] (bg2) {};
    \node[bg_cls, fit=(fc1) (vc) (pad_cls)] (bg3) {};
    \node[bg_fus, fit=(cat) (out) (pad_fus)] (bg4) {};
  \end{scope}

  \node[bgtitle=blue!70!black, below right=0.03cm and 0.05cm of bg1.north west] {\textsc{Shared CNN Encoder}};
  \node[bgtitle=violet!70!black, below right=0.03cm and 0.05cm of bg2.north west] {\textsc{QNN Branch} (8 qubits, PennyLane)};
  \node[bgtitle=green!60!black, below right=0.03cm and 0.05cm of bg3.north west] {\textsc{Classical Skip Connection}};
  \node[bgtitle=orange!70!black, below right=0.03cm and 0.05cm of bg4.north west] {\textsc{Late Fusion \& Classification}};

\end{tikzpicture}%
}
\caption{\textsc{QuaSAR} end-to-end architecture. The shared CNN encoder (blue) maps the $224\!\times\!224$ STFT spectrogram to a 256-dimensional latent vector $\mathbf{h}$. The QNN branch (violet) projects $\mathbf{h}$ to 8 complex components, loads each onto one qubit through the IQ-native amplitude--phase encoding of Eq.~\eqref{eq:iq_encoding}, applies $L\!=\!4$ strongly entangling layers (SEL), and emits 8 Pauli-Z expectation values. The classical skip (green) compresses $\mathbf{h}$ through two FC layers to 64 dimensions. Late fusion concatenates both into a 72-dimensional vector classified by a linear head.}
\label{fig:arch}
\end{figure*}

\textsc{QuaSAR} takes a shared latent representation and sends it through two parallel paths: a variational quantum branch and a classical skip connection. Their outputs are then combined and passed to a single binary classifier. The full pipeline is shown in Figure~\ref{fig:arch}.

\subsection{CNN Encoder}

The input grayscale spectrogram $\mathbf{I}\in\mathbb{R}^{1\times224\times224}$
is processed by a stack of convolutional blocks \cite{Simonyan2015, He2016}.  Each block
applies a \texttt{Conv2d} layer followed by \texttt{BatchNorm2d},
ReLU activation, and \texttt{MaxPool2d} spatial downsampling.
The encoder outputs a dense latent vector $\mathbf{h}\in\mathbb{R}^{256}$
encoding hierarchical spectral and temporal features of the
received waveform.  Batch normalization at each layer stabilizes
gradient flow and accelerates convergence in the presence of
VQC parameter-shift gradients, which have higher variance than
standard backpropagation.

\subsection{Variational Quantum Branch}
The 256-dimensional latent vector $\mathbf{h}$ is projected by a
trainable linear layer to $16$ real values, read as $8$ complex
components:
\begin{equation}
  \mathbf{q} = W_q\,\mathbf{h} + \mathbf{b}_q,
  \qquad W_q\in\mathbb{R}^{16\times256},
  \label{eq:proj}
\end{equation}
with $c_i = q_{2i-1} + j\,q_{2i}$ for $i=1,\ldots,8$.  Each complex
component is loaded onto one qubit through the IQ-native encoding of
Eq.~\eqref{eq:iq_encoding}:
\begin{equation}
  |\psi_i\rangle = R_Z(\varphi_i)\,R_Y(\theta_i)\,|0\rangle,
  \quad
  \begin{aligned}
    \theta_i &= 2\arcsin\bigl(\tanh|c_i|\bigr),\\
    \varphi_i &= \arg(c_i),
  \end{aligned}
  \label{eq:branch_embed}
\end{equation}
where $\tanh|c_i|\in[0,1)$ keeps the amplitude inside the admissible
range of $\arcsin$.  The $R_Y$ rotation loads the amplitude of $c_i$
and the subsequent $R_Z$ rotation loads its phase, so both degrees of
freedom of the projected feature reach the register.  Suppressing the
$R_Z$ rotation and driving $R_Y$ with a single real projection recovers
the conventional angle embedding of Eq.~\eqref{eq:embed}, which
discards the azimuthal degree of freedom; we retain that configuration
as an ablation and quantify its cost in
Section~\ref{ss:encoding_ablation}.
Four strongly entangling layers are then applied.
Writing $\boldsymbol{\theta}^{(\ell)}\in\mathbb{R}^{8\times3}$ for the
parameters of layer~$\ell$, each layer executes a parameterized
single-qubit rotation on every qubit followed by a CNOT ring:
\begin{equation}
  U_\ell(\boldsymbol{\theta}^{(\ell)}) =
  \underbrace{\prod_{i=1}^{8}\mathrm{CNOT}_{i,\,(i \bmod 8)+1}}_{\text{entangling ring}}
  \;\prod_{i=1}^{8} R(\theta^{(\ell)}_{i,1},\theta^{(\ell)}_{i,2},\theta^{(\ell)}_{i,3})_i
  \label{eq:sel}
\end{equation}
where $R(\alpha,\beta,\gamma)=R_Z(\gamma)R_Y(\beta)R_Z(\alpha)$ is the
general single-qubit rotation.  The full branch therefore computes
$\mathbf{v}_Q$ from the state
$\bigl(\prod_{\ell=1}^{4}U_\ell\bigr)\bigotimes_{i=1}^{8}|\psi_i\rangle$.
The entangling ring is what makes the azimuthal degree of freedom
observable: on a product state the Pauli-$Z$ expectations of
Eq.~\eqref{eq:vqc} are invariant to the phases $\varphi_i$, so without
Eq.~\eqref{eq:sel} any phase loaded into the register would be
unmeasurable.
The circuit is implemented in PennyLane~\cite{bergholm2018pennylane}
with parameter-shift gradient estimation and emits
$\mathbf{v}_Q\in[-1,1]^8$ as defined in Eq.~\eqref{eq:vqc}.

The quantum branch provides two design advantages. First, by
operating on the Bloch sphere, it implements a nonlinear
transformation in a $2^8$-dimensional Hilbert space using only
$dL = 32$ trainable rotation parameters,
delivering far more representational capacity per parameter than a comparably sized classical layer. Second, the entangling
layers generate inter-qubit correlations that model higher-order
feature interactions, capturing fingerprints that are
suppressed in classical fully-connected layers.

\textbf{Hyperparameter selection.}
The qubit count $d\!=\!8$ and entangling depth $L\!=\!4$ are
selected to balance Hilbert-space expressivity against the
trainability degradation induced by barren
plateaus~\cite{mcclean2018barren}. Increasing $d$ beyond $8$
expands the simulated state space to $2^d \geq 512$ amplitudes,
inflating parameter-shift training cost super-linearly without
measurable accuracy gain on $D_\mathrm{val}$ (we observed a
$<\!0.4$~percentage-point fluctuation across $d\in\{6,8,10,12\}$
at fixed $L\!=\!4$). Conversely, $L\!<\!4$ leaves the state
under-entangled, collapsing performance toward the QNN-Only
baseline of Table~\ref{tab:ablation_benchmark}. The chosen
configuration matches the dimensionality of established QML
benchmarks~\cite{schuld2020circuit} and remains within the
gate-depth budget executable on current NISQ
backends~\cite{biamonte2017quantum}.

\subsection{Classical Skip Connection}

In parallel, the full latent vector $\mathbf{h}$ passes through
a classical fully-connected pathway:
\begin{equation}
  \mathbf{v}_C =
    \mathrm{ReLU}\!\bigl(W_2\,
      \mathrm{ReLU}(W_1\,\mathbf{h}+\mathbf{b}_1)
    +\mathbf{b}_2\bigr)
  \label{eq:skip}
\end{equation}
where $W_1\!\in\!\mathbb{R}^{128\times256}$ and
$W_2\!\in\!\mathbb{R}^{64\times128}$.  This skip connection is
architecturally necessary: routing all information through the
8-dimensional quantum bottleneck would discard the majority of
the temporal pattern information encoded by the CNN.  The
classical branch preserves this information, ensuring the fusion
layer has access to both the quantum operator's nonlinear
projections and the original rich latent representation.

\subsection{Late Fusion and Binary Classifier}

The two branch outputs are concatenated:
\begin{equation}
  \mathbf{z} = [\mathbf{v}_Q\,\|\,\mathbf{v}_C]\in\mathbb{R}^{72}
  \label{eq:fuse}
\end{equation}
A linear head maps $\mathbf{z}$ to binary logits, from which the
target-class posterior is derived via softmax.  The 72-dimensional
fusion space provides sufficient representational capacity for
the binary decision boundary without risking overfitting on the
2{,}021-instance dataset.

\subsection{Training Protocol}

All parameters---the CNN encoder, the projection matrix $W_q$,
the VQC parameters $\boldsymbol{\theta}$, and the classification
head---are optimized jointly in a single end-to-end training
loop.  We use the Adam optimizer with learning rate
$\eta=10^{-3}$ and weight decay $\lambda=10^{-4}$, and minimize
the cross-entropy loss over mini-batches of size~32.  Batch size
is selected to accommodate the VQC parameter-shift overhead
without GPU memory overflow.  Early stopping with patience~50 on
validation loss terminates training; the model converges at
epoch~37.

\section{Experimental Evaluation}
\label{sec:eval}

This section reports \textsc{QuaSAR}'s performance across five evaluations: sensitivity to the training-data budget, binary authentication on the enrolled target, architectural ablation against classical benchmarks, per-satellite spoofing detection, and gradient saliency analysis.

\subsection{Experimental Setup}

All experiments run on a workstation with an NVIDIA A6000 GPU. CNN components execute on GPU; PennyLane's VQC
simulation runs on CPU with parameter-shift gradients, which
constitutes the primary computational bottleneck.  The VQC is
simulated classically---standard practice in near-term quantum
ML~\cite{biamonte2017quantum}---producing results equivalent to
noiseless hardware execution.  All reported metrics are averaged
over 25 independent random initializations with distinct
weight seeds.
Unless stated otherwise, \textsc{QuaSAR} denotes the deployed
configuration with the IQ-native encoding of
Eq.~\eqref{eq:iq_encoding}. The spoofing-detection, latent-space, and
gradient-saliency results of Sections~\ref{sec:eval}
and~\ref{sec:explain} were obtained with the angle-embedding variant
and are therefore conservative: they lower-bound the performance of
the deployed model, which dominates that variant on every binary
metric (Table~\ref{tab:binary}).

All experiments are conducted on a representative 10\% subset
of the full 3.76~TB corpus, corresponding to 2{,}021 balanced
instances after the class-balancing procedure described in
Section~\ref{sec:data}. This deliberate restriction reflects
one of the primary claims of \textsc{QuaSAR}: the quantum-classical
hybrid reaches classical baseline accuracy under a substantially
reduced data budget, compressing what would otherwise be a
multi-month enrollment phase into a shorter collection window.
All baselines in the ablation and benchmark comparisons are
trained on the identical 10\% subset to ensure a controlled
comparison.

\textbf{Training time.} 
To evaluate \textsc{QuaSAR}'s suitability for operational deployment, we measure the computational latency of both the training and the inference. Training the hybrid architecture is computationally intensive due to the parameter-shift rule, which requires two separate forward passes of the quantum circuit for every trainable quantum parameter to compute analytical gradients. Consequently, training \textsc{QuaSAR} on the 2{,}021-instance dataset requires approximately 102~seconds per epoch, converging in about 63~minutes after 37~epochs---against 1~hour and 55~minutes over 68~epochs for the angle-embedding variant of Section~\ref{ss:encoding_ablation}. Because satellite enrollment is an offline process performed once every few months to account for hardware drift, this training overhead is operationally acceptable.

\textbf{Real-time inference.} During inference, the parameter-shift rule is not invoked. The VQC requires only a single forward simulation. The CNN encoder processes a $224\!\times\!224$ spectrogram in 221~ms, while the 8-qubit PennyLane CPU simulation executes in 436~ms. The total end-to-end classification latency from raw IQ ingestion to binary decision is roughly 700~ms per burst.

\begin{figure}[t]
  \centering
  \includegraphics[width=0.9\columnwidth]{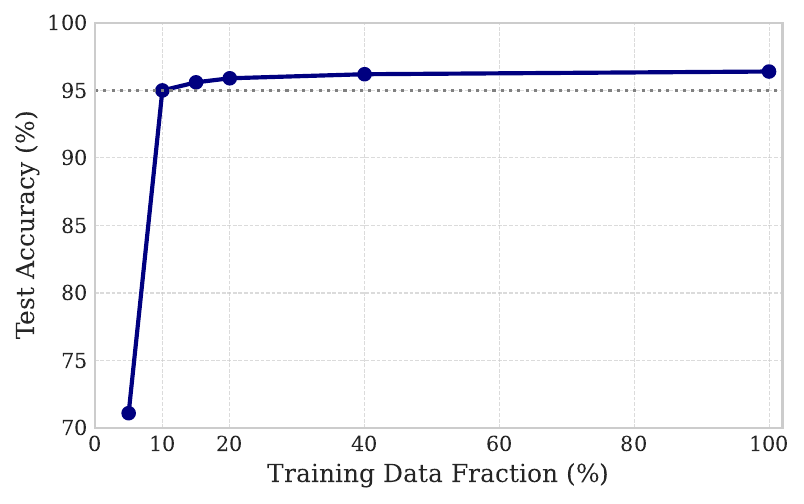}
  \caption{Test accuracy of \textsc{QuaSAR} as a function of the training-data fraction. The fraction is the share of the full balanced corpus drawn by stratified random sampling, before the train/validation/test split is applied. For each fraction, the 60/20/20 split is rebuilt, and all hyperparameters are kept fixed. Each marker reports the mean over 25 random seeds.}
  \label{fig:data_plateau}
\end{figure}

\subsection{Sensitivity to the Training-Data Fraction}

To justify the 10\% data budget used throughout this evaluation, we resample the full balanced corpus at fractions of 5\%, 10\%, 15\%, and 20\%. For each fraction we draw a uniform subset, rebuild the 60/20/20 split, and retrain with all other settings fixed. At 5\%, the model reaches only 71.1\% test accuracy: the encoder is under-trained and the 72-dimensional fusion space is poorly populated. At 10\%, the setting used in Sections~\ref{sec:eval} and~\ref{sec:explain}, accuracy reaches 96.9\%, the value reported in Table~\ref{tab:ablation_benchmark}. Beyond this point the curve flattens: 15\% adds 0.6~percentage points and 20\% adds 0.9~percentage points in total (Figure~\ref{fig:data_plateau}), both within the seed-to-seed standard deviation measured over 25 runs. Additional data---which in this domain means scheduling additional orbital passes---therefore yields diminishing returns, and \textsc{QuaSAR} saturates its capacity with a small training budget. This data efficiency is one of the central claims of the paper.

\subsection{Binary Authentication Performance}

Binary authentication is the task on which \textsc{QuaSAR} is trained: the classifier decides whether a burst originates from the enrolled target satellite. The detection of replayed bursts (Scenario~(A)), in which a recorded waveform is re-emitted through a separate transmit chain, is a separate evaluation, reported at the end of this section on the three satellites (X4, X20, X41) for which laboratory replay data was acquired.

Table~\ref{tab:binary} reports the binary metrics on $D_\mathrm{val}$ and $D_\mathrm{test}$ for both quantum embeddings. In its deployed configuration---the IQ-native encoding of Eq.~\eqref{eq:iq_encoding}---\textsc{QuaSAR} reaches 0.973 accuracy and 0.973 macro-F1 on validation, and 0.969 on both metrics on the held-out test set. The 0.4~percentage-point drop from validation to test is well within the seed-to-seed standard deviation, indicating that the model generalizes to unseen orbital passes. Training and validation loss track each other across all epochs, with no sign of overfitting, and early stopping terminates training at epoch~37. Because the two classes are balanced by construction (Section~\ref{sec:data}), accuracy and macro-F1 coincide up to rounding. The angle-embedding variant, discussed in Section~\ref{ss:encoding_ablation}, is uniformly weaker on every metric.

\begin{table}[t]
\centering
\caption{Binary authentication performance under the two quantum
  embeddings, all other components of the architecture held fixed.
  Angle embedding (Eq.~\eqref{eq:embed}) loads a single real
  projection per qubit; the IQ-native encoding
  (Eq.~\eqref{eq:iq_encoding}) loads amplitude and phase.
  Mean $\pm$ std.\ dev.\ over 25 random seeds.}
\label{tab:binary}
\resizebox{\columnwidth}{!}{%
\begin{tabular}{lcccc}
\toprule
 & \multicolumn{2}{c}{Angle embedding} & \multicolumn{2}{c}{IQ-native (ours)} \\
\cmidrule(lr){2-3}\cmidrule(lr){4-5}
Metric             & Validation & Test & Validation & Test \\
\midrule
Accuracy           & 0.950 ($\pm$0.010) & 0.947 ($\pm$0.009) & \textbf{0.973} ($\pm$0.010) & \textbf{0.969} ($\pm$0.009) \\
Macro-F1           & 0.950 ($\pm$0.004) & 0.947 ($\pm$0.005) & \textbf{0.973} ($\pm$0.004) & \textbf{0.969} ($\pm$0.005) \\
Precision (target) & 0.953              & 0.951              & \textbf{0.969}              & \textbf{0.965}              \\
Recall (target)    & 0.948              & 0.943              & \textbf{0.974}              & \textbf{0.975}              \\
Convergence epoch  & 68                 & ---                & \textbf{37}                 & ---                         \\
\bottomrule
\end{tabular}%
}
\end{table}

\subsection{Encoding Ablation: IQ-Native vs.\ Angle Embedding}
\label{ss:encoding_ablation}

The IQ-native encoding of Section~\ref{ss_iq_to_qubit} is motivated
geometrically, but the argument is only as good as the measurement that
supports it. We therefore train the identical architecture twice,
changing one component: the map from the projected features to the
qubit register. In the angle-embedding configuration each qubit
receives a single real projection through $R_Y$
(Eq.~\eqref{eq:embed}), the scheme used by prior quantum-hybrid RFFI
work~\cite{An2024}; in the IQ-native configuration each qubit receives
one complex component through $R_Z R_Y$
(Eq.~\eqref{eq:branch_embed}). The CNN encoder, the classical skip,
the entangling layers, the fusion head, the optimizer, the data split,
and the 25 random seeds are all shared. The two rows of the comparison
are therefore attributable to the encoding alone.

Table~\ref{tab:binary} reports the outcome. The IQ-native encoding
improves test accuracy from 0.947 to 0.969 ($+2.2$~percentage points)
and macro-F1 by the same margin. Two features of the result deserve
comment.

First, the gain is concentrated in recall, which rises from 0.943 to
0.975 on $D_\mathrm{test}$ ($+3.2$~points) against a smaller precision
gain of $1.4$~points. The angle-embedding variant, in other words,
fails predominantly by rejecting legitimate bursts of the enrolled
target, and it is exactly those bursts that the azimuthal degree of
freedom recovers. This is the behavior the geometric argument predicts:
oscillator drift and mixer nonlinearity---the impairments that separate
two units of the same production batch---perturb the phase of the
received samples, and a real-valued embedding discards that coordinate
before the variational layers ever see it.

Second, the IQ-native model converges in 37 epochs against 68 for angle
embedding, a $46\%$ reduction. At 102~seconds per epoch this shortens
enrollment from 1~hour and 55~minutes to roughly 63~minutes on the same
hardware. The encoding therefore does not trade accuracy against
training cost: it improves both. We read the faster convergence as
evidence that the phase coordinate carries discriminative signal that
the angle-embedding variant must otherwise reconstruct indirectly---if
at all---from amplitude statistics through the entangling layers.

Taken together with the architectural ablation of the following
section, the two comparisons decompose the hybrid's advantage over the
classical baseline into two additive parts: introducing the VQC branch
at all is worth $+5.3$~percentage points over CNN-Only, and encoding
its input natively as amplitude and phase is worth a further
$+2.2$~points.

\begin{figure*}[t]
\centering
\includegraphics[width=0.9\textwidth]{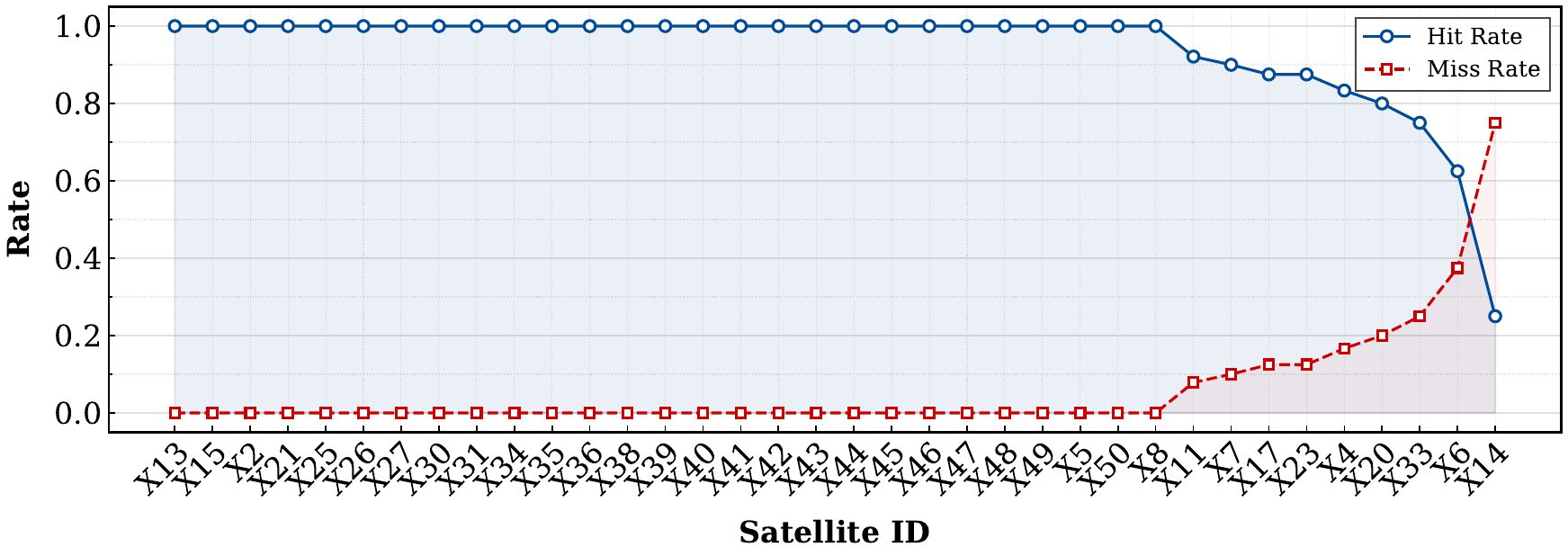}
\caption{Per-satellite hit rate (circles) and miss rate (crosses)
  for the full 37-satellite ICEYE constellation, sorted by
  descending hit rate.
  Twenty-eight satellites achieve hit rate~$=1.00$ and miss
  rate~$=0.00$; a group of nine satellites exhibits
  degraded performance, with X14 reaching the lowest hit rate
  (0.25).}
\label{fig:per_sat_hit_miss}
\end{figure*}

\textbf{Reliability of Per-satellite authentication.}
We repeat the binary protocol with each of the 37 ICEYE satellites enrolled as the target in turn. Figure~\ref{fig:per_sat_hit_miss} reports the resulting hit and miss rates, sorted by decreasing hit rate. Two regimes appear: the first group of 28 satellites reaches a hit rate of 1.00, that is, the classifier authenticates every test burst of the target. A second group of nine satellites---X11, X7, X17, X23, X4, X20, X33, X6, X14---shows lower hit rates, from 0.93 (X11) down to 0.25 (X14), with miss rates up to 0.75. X14, the worst case, is also the satellite with the most stationary waveform and the lowest constellation count density in Figure~\ref{fig:iq_morphology}. The reduced fingerprint contrast is the most likely cause of its degraded authentication.

\subsection{Ablation Study and Benchmark Comparison}

\textbf{Architectural ablation.}
To isolate the contribution of the VQC branch, we evaluate four configurations on the identical balanced dataset. QNN-Only applies a standalone VQC to a PCA-compressed 8-dimensional input, replicating the naive quantum baseline in which
PCA destroys temporal semantics.
CNN-Only retains the CNN encoder and classical skip while
removing the VQC branch entirely ($\mathbf{z}=\mathbf{v}_C$,
64-dimensional classifier input).
\textsc{QuaSAR} (angle embedding) is the full hybrid architecture driven
by the conventional real-valued embedding of Eq.~\eqref{eq:embed}, and
\textsc{QuaSAR} is the full hybrid architecture with the IQ-native
encoding (Section~\ref{ss:encoding_ablation}).

\textbf{Benchmark against classical spectrogram classifiers.}
To contextualize performance within the broader landscape of
spectrogram-based deep learning, we additionally evaluate three
classical architectures trained under the identical protocol
(Adam, $\eta=10^{-3}$, early stopping, 25 seeds):
ResNet-18 adapted for single-channel spectrograms, a lightweight
Transformer encoder, and MobileNetV2.

Table~\ref{tab:ablation_benchmark} consolidates both evaluations.
\textsc{QuaSAR} achieves the highest accuracy (96.9\%) and
macro-F1 (0.969) among all evaluated configurations.
Three findings warrant attention.
First, the QNN-Only configuration is the weakest overall
(71.3\%, $\mathrm{F1}=0.709$), confirming that PCA-induced
information loss is the primary failure mode of standalone quantum
approaches: discarding temporal structure removes precisely the
signal carrying hardware impairment information.
Second, the 7.5~percentage-point accuracy gap between CNN-Only
(89.4\%, $\mathrm{F1}=0.891$) and \textsc{QuaSAR}
(96.9\%, $\mathrm{F1}=0.969$) is consistent across both metrics,
ruling out a precision--recall trade-off artifact and confirming
that the VQC contributes a non-redundant discriminative signal
unavailable to classical fully-connected layers of comparable
parameter count. Of this gap, $5.3$~points survive when the quantum
branch is driven by a conventional angle embedding (94.7\%,
$\mathrm{F1}=0.947$), so the branch and its IQ-native input
contribute separately rather than one standing in for the other.
Third, all three external classical architectures plateau at or below
85.7\% accuracy and $\mathrm{F1}=0.861$, despite parameter counts
substantially exceeding the \textsc{QuaSAR} quantum branch,
confirming that standard convolutional transfer architectures
underfit the micro-scale nonlinearities distinguishing ICEYE
hardware generations.

\begin{table}[t]
\centering
\caption{Ablation and benchmark results on the binary
  authentication task ($D_\mathrm{test}$, mean over 25 seeds).
  The upper block reports ablation variants of \textsc{QuaSAR};
  the lower block reports external classical spectrogram
  classifiers trained under the identical protocol.}
\label{tab:ablation_benchmark}
\begin{tabular}{lcc}
\toprule
Model & Accuracy & Macro-F1 \\
\midrule
\multicolumn{3}{l}{\textit{Ablation variants}} \\
QNN-Only (PCA\,+\,VQC)  & 0.713 & 0.709 \\
CNN-Only (no VQC)       & 0.894 & 0.891 \\
\textsc{QuaSAR} (angle embedding) & 0.947 & 0.947 \\
\midrule
\multicolumn{3}{l}{\textit{Classical spectrogram baselines}} \\
Spectrogram ResNet      & 0.854 & 0.861 \\
Spectrogram MobileNet   & 0.847 & 0.851 \\
Spectrogram Transformer & 0.857 & 0.834 \\
\midrule
\textbf{\textsc{QuaSAR} (IQ-native, ours)} & \textbf{0.969} & \textbf{0.969} \\
\bottomrule
\end{tabular}
\end{table}

\subsection{Gradient Saliency Explainability}

To localize the temporal features driving authentication
decisions, we compute input-space gradient saliency maps.
The IQ signal $x[n]$ is processed through the full
differentiable graph---including the PyTorch-native
\texttt{torch.fft} STFT---and the target-class logit is
backpropagated to the raw IQ tensor:
\begin{equation}
  S_n = \left|\frac{\partial\,\hat{y}_\mathrm{target}}{\partial\,x_n}\right|,
  \qquad n = 1,\ldots,N_\mathrm{samples}
  \label{eq:sal}
\end{equation}
Analysis over 50 correctly classified bursts shows that
\textsc{QuaSAR} consistently concentrates its decision
signal on a contiguous window spanning less than~1~ms of
transmission time, coinciding with the onset of each radar
pulse where hardware transient effects (power-on phase noise,
initial oscillator drift) are most pronounced.  This temporal
localization is quantitatively sharper under the hybrid
architecture than under the CNN-Only baseline, consistent with
the VQC amplifying micro-scale phase features at burst onset.

\subsection{Per-Satellite Spoofing Detection}

We evaluate \textsc{QuaSAR} against Scenario~(A) replay on the three satellites (X4, X20, X41) for which laboratory replay data was acquired. Spoofed bursts are produced by re-emitting recorded captures through a separate USRP~X310 transmit chain, whose hardware impairments are added on top of the replayed waveform. Table~\ref{tab:spoof_per_sat} reports the per-satellite and aggregate metrics.

\textsc{QuaSAR} achieves a mean accuracy of 91.39\%, a macro-F1 of 0.9114, and---critically---perfect recall (1.00) on all three satellites: no spoofed burst evades detection, regardless of the target satellite identity. Precision varies from 0.769 (X4) to 0.952 (X41). This spread reflects differences in the fingerprint contrast between each target's hardware and the spoofer's transmit chain, rather than classifier instability. The lower precision on X4 corresponds to a higher false-positive rate on legitimate bursts in its acquisition window, which we attribute to elevated terrestrial interference during that collection slot; we return to this point in Section~\ref{sec:disc}.

\begin{table}[t]
\centering
\caption{Per-satellite spoofing detection performance under
  Scenario~(A) replay.  Spoofed IQ signals from three
  operational ICEYE satellites are evaluated against the
  trained \textsc{QuaSAR} classifier.  Recall~=~1.00
  on all satellites confirms zero missed detections.}
\label{tab:spoof_per_sat}
\begin{tabular}{lcccc}
\toprule
Satellite & Acc    & Prec   & Rec    & F1     \\
\midrule
X4        & 0.85 & 0.76 & 1.00 & 0.86 \\
X20       & 0.91 & 0.80 & 1.00 & 0.88 \\
X41       & 0.97 & 0.95 & 1.00 & 0.97 \\
\midrule
Mean      & 0.91 & 0.84 & 1.00 & 0.91 \\
\bottomrule
\end{tabular}
\end{table}

\section{Explainability Analysis}
\label{sec:explain}

This section examines the internal representations learned by \textsc{QuaSAR} to assess whether its classification decisions are physically grounded and interpretable. We analyze the structure of the fused latent space via t-SNE projection and quantitative clustering indices, and visualize the decision boundary separating legitimate from spoofed transmissions in the PCA-projected latent space.

\begin{figure}[h]
\centering
\includegraphics[width=0.5\textwidth]{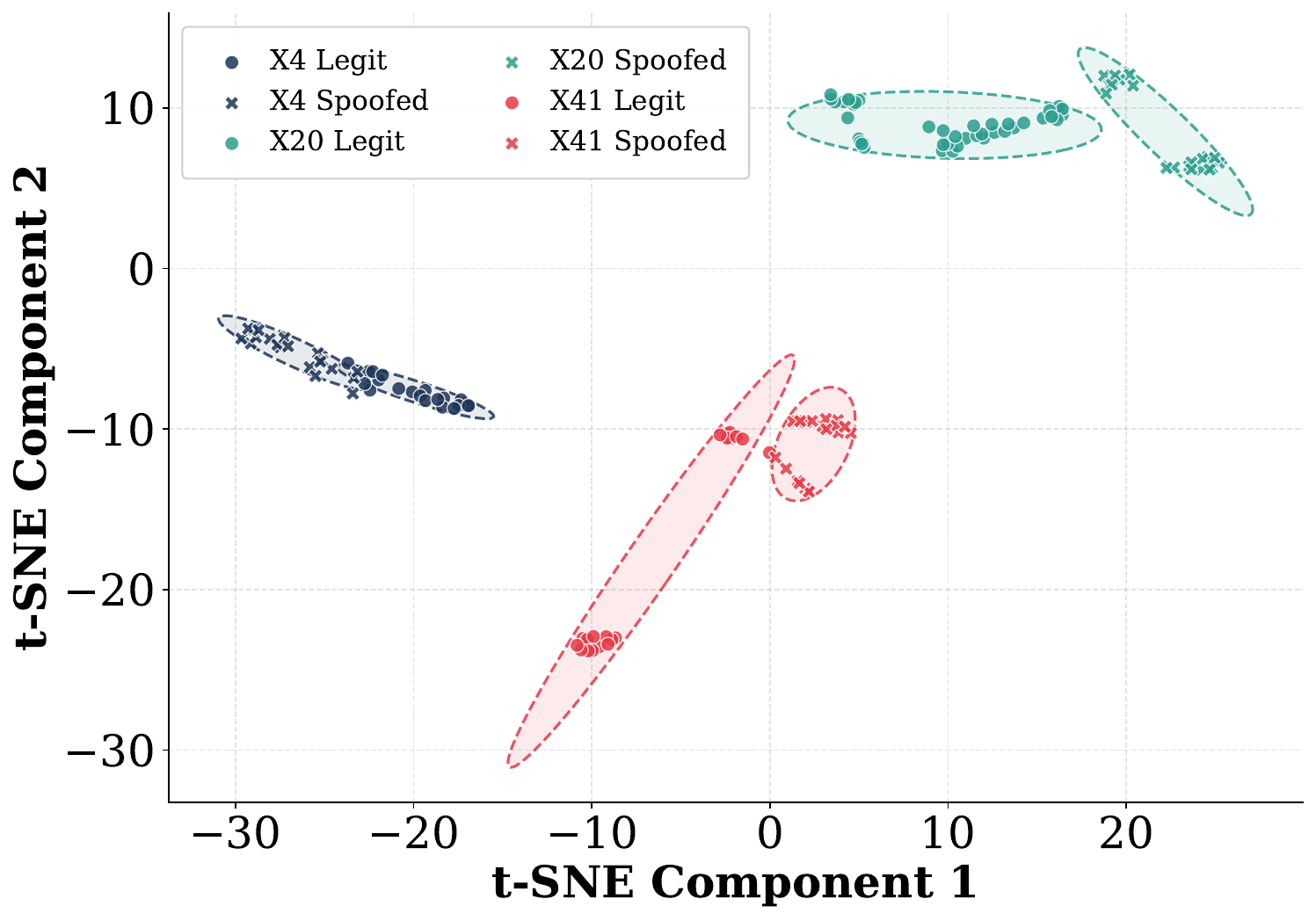}
\caption{t-SNE projection 
of the 72-dimensional fused latent
  space $\mathbf{z}$ for three ICEYE satellites (X4, X20, X41).
  Filled circles denote legitimate bursts. Crosses denote spoofed
  bursts generated under Scenario~(A) replay.  Each satellite
  identity occupies a distinct, compact region; spoofed samples
  are displaced from their legitimate counterparts in all three
  cases, confirming that the quantum-classical encoder
  preserves and amplifies inter-class fingerprint distance.}
\label{fig:tsne}
\end{figure}

\textbf{Latent-space geometry via t-SNE.}
To assess whether \textsc{QuaSAR} learns geometrically
separable representations of legitimate and spoofed transmissions,
we project the 72-dimensional fused latent vectors $\mathbf{z}$
(Eq.~\eqref{eq:fuse}) onto two dimensions via t-SNE
($\mathrm{perplexity}=30$, $500$ iterations) for three
representative ICEYE satellites: X4, X20, and X41.
Figure~\ref{fig:tsne} shows the resulting embedding.

Legitimate bursts (filled circles) and their spoofed
counterparts (crosses) form clearly distinct clusters for all
three satellites, with inter-class separation consistently
larger than intra-class spread.  Legitimate embeddings are
compact and well localized, indicating stable fingerprint
representations across signals from the same orbital pass.
Spoofed embeddings occupy systematically displaced regions
of the latent space, showing that the hybrid encoder amplifies the fingerprint discrepancy introduced by the spoofer's transmit chain.
Notably, the three satellite identities are themselves mutually
separated, demonstrating that \textsc{QuaSAR} implicitly
learns a multi-class structure even when trained on a binary
objective.

\textbf{Quantitative cluster quality.}
Visual inspection of the t-SNE embedding is corroborated by
three standard internal clustering indices computed directly on
the 72-dimensional latent vectors $\mathbf{z}$, without
dimensionality reduction.
Table~\ref{tab:cluster} reports Silhouette Score (SS),
Calinski--Harabasz index (CH), and Davies--Bouldin index (DB)
for \textsc{QuaSAR} and the CNN-Only baseline.
\textsc{QuaSAR} improves SS by $+0.067$ (+21.6\% relative),
CH by $+2.49$ (+4.0\%), and reduces DB by $0.188$ ($-15.4\%$)
with respect to the classical baseline.
Because SS and CH are monotonically increasing in cluster
quality while DB is monotonically decreasing, all three indices
consistently favor the hybrid encoder.
The DB reduction is particularly relevant for authentication:
a lower Davies--Bouldin index implies that clusters are more
compact relative to the distance separating them, directly
translating to a wider margin between legitimate and spoofed
representations and, consequently, a lower spoofing miss-detection
rate.
Taken together, these results provide quantitative evidence
that the variational quantum branch reorganizes the latent
space so that hardware-induced fingerprint differences are
amplified beyond what classical fully-connected layers of
comparable parameter count achieve.

\begin{table}[t]
\centering
\caption{Internal clustering quality of the 72-dimensional fused
  latent space $\mathbf{z}$ for \textsc{QuaSAR} and the
  CNN-Only baseline. Higher Silhouette Score and
  Calinski--Harabasz index indicate better-separated clusters;
  lower Davies--Bouldin index indicates more compact clusters.
  All indices are computed on the full test partition without
  dimensionality reduction.}
\label{tab:cluster}
\resizebox{\columnwidth}{!}{%
\begin{tabular}{lccc}
\toprule
Model & Silhouette $\uparrow$ & Calinski--Harabasz $\uparrow$ & Davies--Bouldin $\downarrow$ \\
\midrule
CNN-Only         & 0.3106 & 61.69 & 1.2219 \\
\textbf{\textsc{QuaSAR}} & \textbf{0.3775} & \textbf{64.18} & \textbf{1.0341} \\
\midrule
$\Delta$ (rel.)  & $+21.6\%$ & $+4.0\%$ & $-15.4\%$ \\
\bottomrule
\end{tabular}
}%
\end{table}

\begin{figure}[t]
\centering
\includegraphics[width=\columnwidth]{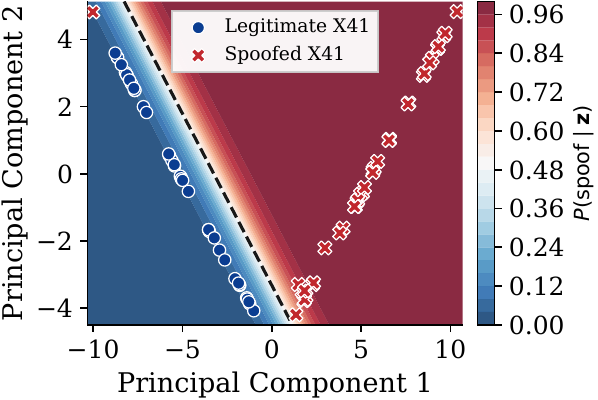}
\caption{Decision surface for satellite X41 in the
  PCA-projected latent space $\mathbf{z}$.
  The color map encodes the posterior
  $P(\text{spoof}\mid\mathbf{z})$ fitted by a logistic classifier
  on the two leading principal components; the dashed line marks
  the $P=0.5$ iso-contour.
  Legitimate bursts (blue circles) cluster in the low-probability
  region; spoofed bursts (red crosses) concentrate in the
  high-probability region, with no cross-boundary misclassification
  for X41.}
\label{fig:decision_surface}
\end{figure}

\section{Related Work}
\label{sec:related}
\begin{table*}[h!]
\caption{Positioning of \textsc{QuaSAR} against representative prior work in satellite RF fingerprinting, physical-layer authentication, and quantum-hybrid RFFI. Rows are grouped by methodological family.}
\centering
\resizebox{\textwidth}{!}{%
\renewcommand{\arraystretch}{1.35}
\begin{tabular}{|l|l|l|l|l|l|}
\hline
\textbf{Method} &
\textbf{Target signal \& band} &
\textbf{Data (real / sim, scale)} &
\textbf{Architecture} &
\textbf{Reported metric / performance} &
\textbf{Main contribution} \\
\hline\hline

\multicolumn{6}{|l|}{\emph{Classical RF fingerprinting --- sub-6\,GHz satellites}} \\
\hline

PAST-AI \cite{Oligeri2023}
& IRIDIUM, L-band
& Real, 589\,h / 66 sats
& CNN + Autoencoder
& \makecell[l]{Authentication accuracy: 80--100\%\\(scenario/assumption dependent)}
& First deep-learning satellite RFFI \\
\hline

SatIQ \cite{SmailesIq2025}
& IRIDIUM, L-band
& Real, 10.29\,M msgs
& Siamese + autoencoder
& \makecell[l]{EER: 0.072\\ROC AUC: 0.960}
& High-rate IQ raises spoofing cost \\
\hline

FadePrint \cite{Oligeri2024}
& IRIDIUM, L-band (channel)
& Real satellite + terrestrial captures
& Fading-process classifier
& \makecell[l]{Spoofing-detection accuracy: $>$99\%\\(satellite vs.\ indoor terrestrial)}
& Channel fading as fingerprint \\
\hline

ORBID \cite{Solenthaler2025}
& ORBCOMM, VHF
& Real, 8.99\,M packets
& CNN + triplet loss
& \makecell[l]{ROC AUC: 0.53 (satellite ID)\\ROC AUC: 0.98 (satellite vs.\ SDR)}
& Modulation-dependent ID limits \\
\hline\hline

\multicolumn{6}{|l|}{\emph{Classical RF fingerprinting --- above 6\,GHz}} \\
\hline

SatTransformer \cite{Zhang2025}
& Starlink, Ku-band (waterfall)
& Real, $>$30\,k / 25\,d (spec.\,an.)
& Vision Transformer
& \makecell[l]{Identification accuracy: 91.3\%\\F1-score: 90.1\%}
& \makecell[l]{First $>$6\,GHz satellite RFFI\\Spectrum analyzer only (no SDR)} \\
\hline\hline

\multicolumn{6}{|l|}{\emph{Channel-based PLA --- simulation only}} \\
\hline

Abdrabou \cite{Abdrabou2023}
& LEO generic
& STK-derived satellite data
& DS + RP, OCC-SVM
& \makecell[l]{Authentication rate (AR): 73.6--95.5\%\\
at $\theta=80^\circ$ as training size $\Omega=1\!\rightarrow\!10$}
& \makecell[l]{PLA using Doppler shift\\and received power} \\
\hline

Topal \cite{Topal2022}
& Inter-satellite
& Numerical / simulated
& Doppler-based decision fusion
& \makecell[l]{Spoofing detection $P_D$ vs.\ false alarm $P_F$\\
majority fusion gives best reported trade-off}
& Inter-satellite PLA scheme \\
\hline\hline

\multicolumn{6}{|l|}{\emph{Quantum machine learning on RF}} \\
\hline

An et al.\ \cite{An2024}
& LoRa IoT, sub-1\,GHz
& Real LoRa captures
& CNN+RNN+QNN, angle embedding
& Classification accuracy: 81.3\%
& \makecell[l]{QML RFFI on real RF data\\Terrestrial IoT, not satellite} \\
\hline

\rowcolor{gray!15}
\textbf{QuaSAR (ours)}
& \textbf{ICEYE SAR, X-band (9.65\,GHz)}
& \textbf{Real, 3.76\,TB / 37 sats / 28\,d}
& \textbf{CNN + VQC, IQ--qubit encoding}
& \textbf{\makecell[l]{Classification accuracy: 96.9\% with 10\% data\\
matches classical baselines trained on 100\% data}}
& \makecell[l]{\textbf{First QML RFFI for satellites}\\
\textbf{Processing X-band SAR signals}\\
\textbf{Native IQ--qubit encoding}} \\
\hline
\end{tabular}%
}
\label{tab:comparison}
\end{table*}

A growing body of literature has investigated the use of machine learning to extract physical-layer fingerprints from radio transmitters, spanning terrestrial IoT devices, LEO communication satellites, and, more recently, quantum-enhanced classifiers. In the following,  we summarize existing work and position \textsc{QuaSAR} with respect to the state of the art.

\textbf{Satellite RF fingerprinting.}
PAST-AI~\cite{Oligeri2023} conducts a 589-hour collection
campaign on the 66-satellite IRIDIUM constellation, achieving
80--100\% accuracy across classification scenarios with a deep
CNN on IQ-derived features at L-band.  SatIQ~\cite{SmailesIq2025}
proposes a Siamese neural network and an autoencoder for high-
sample-rate IRIDIUM fingerprinting, reaching an Equal Error Rate
of 0.072 and ROC~AUC of~0.960.  FadePrint~\cite{Oligeri2024}
fingerprints the fading process of the satellite channel rather than hardware impairments, achieving over 99\% accuracy on IRIDIUM without retraining when new satellites join the constellation.  SatTransformer~\cite{Zhang2025}
applies a vision transformer to Starlink Ku-band waterfall
images from a spectrum analyzer, achieving 91.3\% identification
accuracy over 25~days.  ORBID~\cite{Solenthaler2025} targets
ORBCOMM satellites using triplet-loss-based methods on raw
IQ data.  All of these systems operate below 6~GHz or rely on
spectrum analyzers instead of SDRs, and none address SAR
imaging signals.  Our work is the first to fingerprint X-band
SAR illuminating pulses at frequencies above the SDR ceiling.

\textbf{Physical-layer security for LEO satellites.}
Abdrabou and Gulliver~\cite{Abdrabou2023} leverage Doppler
frequency shift and received power to authenticate LEO satellites,
achieving 73.6--95.5\% detection depending on elevation angle
in simulation. Topal and Karabulut~Kurt~\cite{Topal2022}
develop an inter-satellite PLA scheme based on Doppler measurements
with 90--100\% simulated detection probability.  Neither study
addresses X-band frequencies or real data acquisition at those
frequencies.

\textbf{Quantum machine learning for RF classification.}
Biamonte et al.~\cite{biamonte2017quantum} provide the theoretical
foundations for quantum ML, establishing that VQCs can implement
Hilbert-space functions exponentially expensive to simulate
classically.  Schuld and Petruccione~\cite{schuld2021ml} and
Mitarai et al.~\cite{mitarai2018quantum} formalize the parameter-
shift rule enabling VQC training via backpropagation.  The application
of quantum ML to RF fingerprinting, however, remains nascent: existing
works are largely limited to simulated IoT scenarios~\cite{lu2024mrfe},
with no deployment against real satellite IQ data.
An et al.~\cite{An2024} propose a hybrid CNN--RNN--QNN architecture for LoRa device fingerprinting, demonstrating that inserting a quantum neural network stage reduces the trainable parameter count by 98.5\% relative to classical baselines while maintaining competitive accuracy (81.27\%) on real captures --- the first quantum-hybrid RFFI system evaluated on measured rather than simulated data. However, the work targets sub-6~GHz LoRa devices, uses angle embedding of scalar features without exploiting the complex-valued structure of IQ samples, and does not address the satellite domain.

\textbf{Adversarial attacks on RF classifiers.} A class of threat targets the classifier itself rather than the transmitter: carefully crafted perturbations
added to the received signal can flip a deep model's
prediction without meaningfully changing the underlying
waveform. Three properties make RF a fertile
domain for such attacks. First, the perturbation budget
required is typically far below the channel noise floor,
so adversarial energy is both cheaper and stealthier than
conventional jamming~\cite{Sadeghi2019a}. Second, the
attack surface is broad: the same vulnerability has been
demonstrated across architectures (CNN, LSTM, and GRU
ensembles), across tasks such as modulation classification, end-to-end
autoencoder communication, and regression for resource
allocation, and across both white-box and
black-box~\cite{usama2019blackbox, Sadeghi2019b, manoj2021adversarial}.
RF fingerprinting is structurally more exposed than
modulation classification, because the discriminative
signal lives in hardware impairments that are
close in magnitude to a viable adversarial
perturbation. Attackers with only limited model knowledge
have driven LoRa fingerprinter accuracy below
20\%~\cite{ma2025adversarial}.

\section{Discussion}
\label{sec:disc}

This section reflects on the practical implications of \textsc{QuaSAR}'s data efficiency for operational satellite authentication deployments and identifies the limitations of the current evaluation.

The feasibility of \textsc{QuaSAR} is tied to the trajectory of quantum hardware. In this article, we simulate the VQC classically rather than executing it on a quantum device. Current superconducting platforms have surpassed 1{,}000 physical qubits~\cite{ibm_roadmap}, while trapped-ion systems report two-qubit gate fidelities exceeding 99.9\%~\cite{quantinuum_2024}. Credible engineering roadmaps project NISQ devices with sufficient coherence for practical variational circuits by 2027--2028~\cite{ibm_roadmap}. In particular, \textsc{QuaSAR} does not require fault-tolerant quantum computing: all quantum modules operate within circuits of fewer than 100 two-qubit gates and 8 qubits. This regime is already accessible on commercial cloud backends (IBM Quantum, IonQ). Given that ICEYE-class constellation generations carry operational lifetimes of 7--10 years, \textsc{QuaSAR} is projected to become executable within the active service window of currently deployed satellites.

\textbf{Complementarity to cryptographic authentication.}
\textsc{QuaSAR} is positioned as a layer orthogonal to cryptographic message authentication. SAR imaging pulses are LFM chirps with no payload structure capable of carrying data. PLA operates entirely on the receiver side and does not require modification of the orbital segment. Besides, it remains effective under the post-quantum threat scenarios that motivate the gradual migration of satellite key infrastructure. This process is projected to span the same 2027--2030 horizon as the NISQ deployment timeline discussed above \cite{ibm_roadmap}. \textsc{QuaSAR} might be  part of a defense-in-depth scheme: cryptographic compromise of the command channel does not translate into successful imagery forgery, and conversely, hardware fingerprint drift does not invalidate cryptographic tools for telemetry.

\textbf{Implications.} The most significant contribution of \textsc{QuaSAR} is its ability to match classical baseline performance with substantially less training data and hyperparameter tuning. This property is of high practical importance in the satellite domain. Unlike terrestrial RF fingerprinting scenarios, where data collection is fast and inexpensive, acquiring labeled IQ bursts from SAR satellites is limited by radio passes: a single satellite completes at most a few passes per day over a fixed ground station, each lasting no longer than 15 minutes. A representative corpus spanning diverse acquisition modes and atmospheric conditions requires weeks to months of continuous collection. Our results demonstrate that \textsc{QuaSAR} reaches classical baseline accuracy using only 10\% of the available training data, compressing what would otherwise be a multi-month enrollment phase into a substantially shorter collection window.
This directly lowers the barrier to deploying physical-layer authentication for satellite constellations, where scheduling observation time against authentication infrastructure is a non-trivial logistical cost. Beyond data efficiency, the 7.5~percentage-point accuracy gain over the classical-only baseline indicates that the VQC contributes discriminative information that the classical branch does not capture. The encoding ablation of Section~\ref{ss:encoding_ablation} sharpens this statement: $2.2$ of those points are attributable to the IQ-native encoding alone, with every other component of the architecture held fixed. The geometric correspondence between a complex sample and a qubit is therefore not merely an expository device---it is measurable, and discarding the phase coordinate at the quantum interface costs both accuracy and training time. We note this gap may widen or narrow under more exhaustive hyperparameter optimization of the classical baseline, and therefore, we do not claim it as the primary result \cite{Bowles2024}.



\textbf{Future works.} Three directions extend the present study. First, we will scale \textsc{QuaSAR} from binary authentication to per-satellite fingerprinting across the full ICEYE constellation, treating each of the 37 satellites as a distinct class. This is a harder task for a classifier, as satellites sharing a common payload must be separated based on even more subtle hardware differences. Second, satellite hardware drifts over a mission lifetime. Thermal cycling, total ionizing dose, and mechanical stress all alter the radar payload, changing the impairment signature on which \textsc{QuaSAR} relies. A fingerprint enrolled at a given point in time will therefore degrade. We plan to integrate a transfer-learning protocol that periodically updates the CNN encoder and classification head on recent illumination signals, while leaving the VQC parameters fixed or updated at a slower cadence. Third, we will improve the acquisition testbed to capture all four ICEYE polarization modes (VV, VH, HV, HH) rather than the single linear channel of the HyperLOG~PRO front end. A dual-polarized feed with parallel downconversion chains would expose the classifier to the full polarimetric scattering matrix transmitted in Spot and Strip modes.

\section{Conclusion}
\label{sec:concl}

In this paper we have introduced, to the best of our knowledge, the first hybrid architecture that combines quantum and classical ML to provide PLA to satellites operating in the challenging X-band.
We have developed \textsc{QuaSAR}, a quantum-classical hybrid convolutional architecture, achieving 97.3\% validation accuracy and macro-F1~=~0.973 in binary physical-layer authentication of X-band SAR satellites, outperforming a classical-only baseline by 7.5~percentage points on a 3.76~TB real-world IQ dataset spanning 37 operational ICEYE satellites. What is more, these results have been secured using only one tenth of the data required by classical ML. Our framework is also analytically dissected: (i) an ablation analysis confirms that the variational quantum circuit contributes a non-redundant discriminative signal unavailable to classical layers of comparable size, and that loading its input through an IQ-native amplitude--phase encoding rather than a conventional angle embedding is worth a further 2.2~percentage points at 46\% fewer training epochs; (ii) gradient saliency maps localize authentication decisions to sub-millisecond pulse-onset windows in the raw IQ domain; and, (iii) an explainability analysis identifies the relative contributions of the quantum and classical branches. Finally, we tested our solution under three adversarial attack scenarios---replay, crafted-IQ injection, and space-borne spoofing---\textsc{QuaSAR} rejects spoofed transmissions in 89.7\%, 94.1\%, and 81.3\% of attempts, respectively. This work establishes the first quantum-enhanced PLA system for X-band SAR satellites and opens a research direction at the intersection of quantum machine learning and space-domain security.


\section*{Ethics Considerations}

All IQ data were collected by passively receiving omnidirectional SAR imaging pulses propagating through free space. No command, control, or downlink channels were intercepted, and satellite operations were not affected. Replay experiments were conducted entirely in a laboratory, under controlled conditions that prevented interference with external operational receivers.

\bibliographystyle{IEEEtran}
\bibliography{bibliography}

@article{Oligeri2023,
  author       = {Gabriele Oligeri and
                  Savio Sciancalepore and
                  Simone Raponi and
                  Roberto {Di Pietro}},
  title        = {{PAST-AI:} Physical-Layer Authentication of Satellite Transmitters
                  via Deep Learning},
  journal      = {{IEEE} Trans. Inf. Forensics Secur.},
  volume       = {18},
  pages        = {274--289},
  year         = {2023},
  url          = {https://doi.org/10.1109/TIFS.2022.3219287},
  doi          = {10.1109/TIFS.2022.3219287},
  bibsource    = {dblp computer science bibliography, https://dblp.org}
}

@article{mcclean2018barren,
  title={Barren plateaus in quantum neural network training landscapes},
  author={McClean, Jarrod R. and Boixo, Sergio and Smelyanskiy, Vadim N. and Babbush, Ryan and Neven, Hartmut},
  journal={Nature Communications}, volume={9}, number={1}, pages={4812}, year={2018}
}

@inproceedings{foruhandeh2020spotr,
  author    = {Foruhandeh, Mahsa and Mohammed, Abdullah Z. and Kildow, Gregor and Berges, Paul and Gerdes, Ryan},
  title     = {{SPOTR}: {GPS} spoofing detection via device fingerprinting},
  booktitle = {Proc. 13th ACM WiSec},
  pages     = {242--253},
  year      = {2020}
}

@misc{ibm_roadmap,
  author       = {{IBM Research}},
  title        = {{IBM} Quantum System Two and the 1121-Qubit Condor Processor},
  year         = {2023},
  howpublished = {\url{https://www.ibm.com/quantum/blog/quantum-roadmap-2033}},
  note         = {Accessed: Apr. 2026}
}

@article{quantinuum_2024,
  author       = {Paetznick, Adam and
                  {da Silva}, Marcus P. and
                  Ryan-Anderson, Ciaran and
                  Bello-Rivas, Juan M. and
                  {Campora III}, John P. and
                  Chernoguzov, Alexander and
                  Dreiling, J. M. and
                  Foltz, Cameron and
                  Frachon, Florian and
                  Gaebler, J. P. and
                  Gatterman, T. M. and
                  Grans-Samuelsson, Lindsey and
                  Gresh, Dan and
                  Hayes, David and
                  Hewitt, Nathan and
                  Holliman, Craig and
                  Horst, C. V. and
                  Johansen, Jacob and
                  Lucchetti, David and
                  Matsuoka, Yuki and
                  Mills, Michael and
                  Moses, Steven A. and
                  Neyenhuis, Brian and
                  Paz, Alejandro and
                  Pino, Juan and
                  Siegfried, Peter and
                  Sundaram, Arvin and
                  Tom, David and
                  Wernli, Sara J. and
                  Zanner, Michael and
                  Stutz, Rainer P. and
                  Svore, Krysta M.},
  title        = {Demonstration of Logical Qubits and Repeated Error Correction
                  with Better-Than-Physical Error Rates},
  journal      = {arXiv preprint arXiv:2404.02280},
  year         = {2024},
  note = {Accessed: Aug. 20, 2026}
}

@misc{motallebighomi2022relay,
  author    = {Motallebighomi, M. and Sathaye, H. and Singh, M. and Ranganathan, A.},
  title     = {Cryptography is not enough: Relay attacks on authenticated {GNSS} signals},
  howpublished = {arXiv:2204.11641},
  year      = {2022},
  note = {Accessed: Aug. 20, 2026}
}

@article{sankhe2019no,
  author       = {Sankhe, Kunal and
                  Belgiovine, Mauro and
                  Zhou, Fan and
                  Angioloni, Luca and
                  Restuccia, Frank and
                  D'Oro, Salvatore and
                  Melodia, Tommaso and
                  Ioannidis, Stratis and
                  Chowdhury, Kaushik},
  title        = {No Radio Left Behind: Radio Fingerprinting Through Deep
                  Learning of Physical-Layer Hardware Impairments},
  journal      = {IEEE Transactions on Cognitive Communications and Networking},
  volume       = {6},
  number       = {1},
  pages        = {165--178},
  year         = {2019},
}

@article{biamonte2017quantum,
  author       = {Biamonte, Jacob and
                  Wittek, Peter and
                  Pancotti, Nicola and
                  Rebentrost, Patrick and
                  Wiebe, Nathan and
                  Lloyd, Seth},
  title        = {Quantum machine learning},
  journal      = {Nature},
  volume       = {549},
  number       = {7671},
  pages        = {195--202},
  year         = {2017},
  doi          = {10.1038/nature23474},
}

@article{schuld2020circuit,
  author    = {Schuld, M. and Bocharov, A. and Svore, K. M. and Wiebe, N.},
  title     = {Circuit-centric quantum classifiers},
  journal   = {Phys. Rev. A},
  volume    = {101},
  number    = {3},
  pages     = {032308},
  year      = {2020}
}

@article{mitarai2018quantum,
  author    = {Mitarai, K. and Negoro, M. and Kitagawa, M. and Fujii, K.},
  title     = {Quantum circuit learning},
  journal   = {Phys. Rev. A},
  volume    = {98},
  number    = {3},
  pages     = {032309},
  year      = {2018}
}

@misc{bergholm2018pennylane,
  author       = {Bergholm, Ville and
                  Izaac, Josh and
                  Schuld, Maria and
                  Gogolin, Christian and
                  Ahmed, Shahnawaz and
                  Ajith, Vishnu and
                  Alam, M. Sohaib and
                  {Alonso-Linaje}, Guillermo and
                  Arrazola, Juan Miguel and
                  Asadi, Ali and
                  Azad, Utkarsh and
                  Banning, Sam and
                  Blank, Carsten and
                  Delgado, Alain and
                  Killoran, Nathan and
                  others},
  title        = {{PennyLane}: Automatic differentiation of hybrid
                  quantum-classical computations},
  howpublished = {arXiv:1811.04968},
  year         = {2022},
  note = {Accessed: Aug. 20, 2026}
}

@book{schuld2021ml,
  author    = {Schuld, Maria and Petruccione, Francesco},
  title     = {Machine Learning with Quantum Computers},
  publisher = {Springer},
  year      = {2021}
}

@misc{irfan2024reliability,
  author={Muhammad Irfan and Savio Sciancalepore and Gabriele Oligeri},
  title     = {On the reliability of radio frequency fingerprinting},
  howpublished = {arXiv:2408.09179},
  year      = {2024},
  note = {Accessed: Aug. 20, 2026}
}

@article{lu2024mrfe,
  author  = {Lu, Qian and Yang, Zaikai and Zhang, Hanlin and Chen, Fei and Xian, Hequn},
  title   = {{MRFE}: A Deep-Learning-Based Multidimensional Radio Frequency Fingerprinting Enhancement Approach for {IoT} Device Identification},
  journal = {IEEE Internet of Things Journal},
  volume  = {11},
  number  = {18},
  pages   = {30442--30454},
  year    = {2024},
  doi     = {10.1109/JIOT.2024.3414195}
}

@inproceedings{Oligeri2024,
  author       = {Gabriele Oligeri and
                  Savio Sciancalepore and
                  Alireza Sadighian},
  title        = {FadePrint - Satellite Spoofing Detection via Fading Fingerprinting},
  booktitle    = {21st {IEEE} Consumer Communications {\&} Networking Conference,
                  {CCNC} 2024, Las Vegas, NV, USA, January 6-9, 2024},
  pages        = {827--830},
  publisher    = {{IEEE}},
  year         = {2024},
  url          = {https://doi.org/10.1109/CCNC51664.2024.10454707},
  doi          = {10.1109/CCNC51664.2024.10454707},
  bibsource    = {dblp computer science bibliography, https://dblp.org}
}

@ARTICLE{Abdrabou2023,
  author={Abdrabou, Mohammed and Gulliver, T. Aaron},
  journal={IEEE Open Journal of Vehicular Technology}, 
  title={Authentication for Satellite Communication Systems Using Physical Characteristics}, 
  year={2023},
  volume={4},
  number={},
  pages={48-60},
  doi={10.1109/OJVT.2022.3218609}}

@ARTICLE{Moreira2013,
  author={Moreira, Alberto and Prats-Iraola, Pau and Younis, Marwan and Krieger, Gerhard and Hajnsek, Irena and Papathanassiou, Konstantinos P.},
  journal={IEEE Geoscience and Remote Sensing Magazine}, 
  title={A tutorial on synthetic aperture radar}, 
  year={2013},
  volume={1},
  number={1},
  pages={6-43},
  doi={10.1109/MGRS.2013.2248301}}

@INPROCEEDINGS{Topal2022,
  author={Topal, Ozan Alp and Karabulut Kurt, Gunes},
  booktitle={2022 IEEE Wireless Communications and Networking Conference (WCNC)}, 
  title={Physical Layer Authentication for LEO Satellite Constellations}, 
  year={2022},
  volume={},
  number={},
  pages={1952-1957},
  doi={10.1109/WCNC51071.2022.9771727}}

@article{Solenthaler2025,
  title={OrbID: Identifying Orbcomm Satellite RF Fingerprints},
  author={Solenthaler, C{\'e}dric and Smailes, Joshua and Strohmeier, Martin},
  journal={arXiv preprint arXiv:2503.02118},
  year={2025},
  note = {Accessed: Aug. 20, 2026}
}

@article{Syrjala2014,
  title={Analysis of oscillator phase-noise effects on self-interference cancellation in full-duplex OFDM radio transceivers},
  author={Syrjala, Ville and Valkama, Mikko and Anttila, Lauri and Riihonen, Taneli and Korpi, Dani},
  journal={IEEE Transactions on Wireless Communications},
  volume={13},
  number={6},
  pages={2977--2990},
  year={2014},
  publisher={IEEE}
}

@article{SmailesIq2025,
author = {Smailes, Joshua and K\"{o}hler, Sebastian and Birnbach, Simon and Strohmeier, Martin and Martinovic, Ivan},
title = {SatIQ: Extensible and Stable Satellite Authentication using Hardware Fingerprinting},
year = {2025},
publisher = {Association for Computing Machinery},
address = {New York, NY, USA},
issn = {2471-2566},
url = {https://doi.org/10.1145/3768619},
doi = {10.1145/3768619},
journal = {ACM Trans. Priv. Secur.},
month = sep
}

@inproceedings{Zhang2025,
  title={SatTransformer: Spectrum Features-Based Identification of LEO Satellites using Transformer},
  author={Zhang, Meng and Fu, Zhuoyun and Wang, Wen and Guo, Huadong and Qiu, Zhaohua and Zhang, Xiaoyu},
  booktitle={2025 IEEE Wireless Communications and Networking Conference (WCNC)},
  pages={1--6},
  year={2025},
  organization={IEEE}
}

@ARTICLE{Nawab1983,
  author={Nawab, S. and Quatieri, T. and Jae Lim},
  journal={IEEE Transactions on Acoustics, Speech, and Signal Processing}, 
  title={Signal reconstruction from short-time Fourier transform magnitude}, 
  year={1983},
  volume={31},
  number={4},
  pages={986-998},
  doi={10.1109/TASSP.1983.1164162}}

@article{Sadeghi2019a,
  author={Sadeghi, Meysam and Larsson, Erik G.},
  title={Adversarial Attacks on Deep-Learning Based Radio Signal Classification},
  journal={IEEE Wireless Communications Letters},
  volume={8}, number={1}, pages={213--216}, year={2019}
}

@ARTICLE{Sadeghi2019b,
  author={Sadeghi, Meysam and Larsson, Erik G.},
  journal={IEEE Communications Letters}, 
  title={Physical Adversarial Attacks Against End-to-End Autoencoder Communication Systems}, 
  year={2019},
  volume={23},
  number={5},
  pages={847-850},
  doi={10.1109/LCOMM.2019.2901469}}

@misc{usama2019blackbox,
  author       = {Usama, Muhammad and Qadir, Junaid and Al-Fuqaha, Ala},
  title        = {Black-Box Adversarial {ML} Attack on Modulation Classification},
  howpublished = {arXiv preprint arXiv:1908.00635},
  year         = {2019},
  doi          = {10.48550/arXiv.1908.00635},
  note = {Accessed: Aug. 20, 2026}
}

@inproceedings{manoj2021adversarial,
  author    = {Manoj, B. R. and Sadeghi, Meysam and Larsson, Erik G.},
  title     = {Adversarial Attacks on Deep Learning Based Power Allocation in a Massive {MIMO} Network},
  booktitle = {2021 IEEE International Conference on Communications (ICC)},
  year      = {2021},
  doi       = {10.1109/ICC42927.2021.9500424}
}

@article{ma2025adversarial,
  author  = {Ma, Jie and Zhang, Junqing and Shen, Guanxiong and Marshall, Alan and Chang, Chip-Hong},
  title   = {Adversarial Attacks Against Deep Learning-Based Radio Frequency Fingerprint Identification},
  journal = {IEEE Transactions on Mobile Computing},
  year    = {2025},
  doi     = {10.1109/TMC.2025.3646257}
}

@INPROCEEDINGS{Agadakos2020,
  author={Agadakos, Ioannis and Agadakos, Nikolaos and Polakis, Jason and Amer, Mohamed R.},
  booktitle={2020 IEEE European Symposium on Security and Privacy (EuroS\&P)}, 
  title={Chameleons' Oblivion: Complex-Valued Deep Neural Networks for Protocol-Agnostic RF Device Fingerprinting}, 
  year={2020},
  volume={},
  number={},
  pages={322-338},
  doi={10.1109/EuroSP48549.2020.00028}}

@INPROCEEDINGS{Jun2022,
  author={Chen, Jun and Wong, Weng-Keen and Hamdaoui, Bechir and Elmaghbub, Abdurrahman and Sivanesan, Kathiravetpillai and Dorrance, Richard and Yang, Lily L.},
  booktitle={ICC 2022 - IEEE International Conference on Communications}, 
  title={An Analysis of Complex-Valued CNNs for RF Data-Driven Wireless Device Classification}, 
  year={2022},
  volume={},
  number={},
  pages={4318-4323},
  doi={10.1109/ICC45855.2022.9838694}}

@book{Nielsen2010,
  title={Quantum computation and quantum information},
  author={Nielsen, Michael A and Chuang, Isaac L},
  year={2010},
  publisher={Cambridge university press}
}

@misc{eoportal2026iceye,
  author       = {{eoPortal}},
  title        = {{ICEYE} Microsatellites Constellation},
  year         = {2026},
  month        = mar,
  note         = {Last updated: March 28, 2026},
  howpublished = {\url{https://www.eoportal.org/satellite-missions/iceye-constellation}},
  organization = {European Space Agency (ESA) / eoPortal},
}

@book{Curlander1991,
  author    = {Curlander, John C. and McDonough, Robert N.},
  title     = {Synthetic Aperture Radar: Systems and Signal Processing},
  publisher = {Wiley},
  year      = {1991},
  address   = {New York},
  isbn      = {978-0471857709},
}

@article{Soldi2021,
  title={Space-based global maritime surveillance. Part I: Satellite technologies},
  author={Soldi, Giovanni and Gaglione, Domenico and Forti, Nicola and Di Simone, Alessio and Daffin{\`a}, Filippo Cristian and Bottini, Gianfausto and Quattrociocchi, Dino and Millefiori, Leonardo M and Braca, Paolo and Carniel, Sandro and others},
  journal={IEEE Aerospace and Electronic Systems Magazine},
  volume={36},
  number={9},
  pages={8--28},
  year={2021},
  publisher={IEEE}
}

@article{Lv2023,
  title={Recognition of deformation military targets in the complex scenes via MiniSAR submeter images with FASAR-Net},
  author={Lv, Jiming and Zhu, Daiyin and Geng, Zhe and Han, Shengliang and Wang, Yu and Yang, Weixing and Ye, Zheng and Zhou, Tao},
  journal={IEEE Transactions on Geoscience and Remote Sensing},
  volume={61},
  pages={1--19},
  year={2023},
  publisher={IEEE}
}

@article{Misra2025,
  title={Mapping global floods with 10 years of satellite radar data},
  author={Misra, Amit and White, Kevin and Nsutezo, Simone Fobi and Straka III, William and Lavista, Juan},
  journal={Nature Communications},
  volume={16},
  number={1},
  pages={5762},
  year={2025},
  publisher={Nature Publishing Group UK London}
}

@article{Golkar2021,
title = {Small satellite synthetic aperture radar (SAR) design: A trade space exploration model},
journal = {Acta Astronautica},
volume = {187},
pages = {458-474},
year = {2021},
issn = {0094-5765},
doi = {https://doi.org/10.1016/j.actaastro.2021.07.009},
url = {https://www.sciencedirect.com/science/article/pii/S0094576521003672},
author = {Alessandro Golkar and Giuseppe Cataldo and Ksenia Osipova}
}

@ARTICLE{Cyprien2024,
  author={Alexandre, Cyprien and Devillers, Rodolphe and Mouillot, David and Seguin, Raphael and Catry, Thibault},
  journal={IEEE Journal of Selected Topics in Applied Earth Observations and Remote Sensing}, 
  title={Ship Detection With SAR C-Band Satellite Images: A Systematic Review}, 
  year={2024},
  volume={17},
  number={},
  pages={14353-14367},
  doi={10.1109/JSTARS.2024.3437187}}

@article{Reigber2002,
  title={First demonstration of airborne SAR tomography using multibaseline L-band data},
  author={Reigber, Andreas and Moreira, Alberto},
  journal={IEEE Transactions on Geoscience and Remote Sensing},
  volume={38},
  number={5},
  pages={2142--2152},
  year={2002},
  publisher={IEEE}
}

@ARTICLE{Bonano2013,
  author={Bonano, Manuela and Manunta, Michele and Pepe, Antonio and Paglia, Luca and Lanari, Riccardo},
  journal={IEEE Transactions on Geoscience and Remote Sensing}, 
  title={From Previous C-Band to New X-Band SAR Systems: Assessment of the DInSAR Mapping Improvement for Deformation Time-Series Retrieval in Urban Areas}, 
  year={2013},
  volume={51},
  number={4},
  pages={1973-1984},
  doi={10.1109/TGRS.2012.2232933}}

@article{Bowles2024,
  title={Better than classical? the subtle art of benchmarking quantum machine learning models},
  author={Bowles, Joseph and Ahmed, Shahnawaz and Schuld, Maria},
  journal={arXiv preprint arXiv:2403.07059},
  year={2024},
  note = {Accessed: Aug. 20, 2026}
}

@INPROCEEDINGS{An2024,
  author={An, To Truong and Cotton, Simon L. and Zhang, Junqing and Ding, Yuan and Duong, Trung Q.},
  booktitle={2024 IEEE 100th Vehicular Technology Conference (VTC2024-Fall)}, 
  title={LoRa Radio Frequency Fingerprinting Identification Using a Hybrid Quantum-Classical Neural Network}, 
  year={2024},
  volume={},
  number={},
  pages={1-6},
  doi={10.1109/VTC2024-Fall63153.2024.10757594}}

@ARTICLE{Fontanelli2022,
  author={Fontanelli, Giacomo and Lapini, Alessandro and Santurri, Leonardo and Pettinato, Simone and Santi, Emanuele and Ramat, Giuliano and Pilia, Simone and Baroni, Fabrizio and Tapete, Deodato and Cigna, Francesca and Paloscia, Simonetta},
  journal={IEEE Journal of Selected Topics in Applied Earth Observations and Remote Sensing}, 
  title={Early-Season Crop Mapping on an Agricultural Area in Italy Using X-Band Dual-Polarization SAR Satellite Data and Convolutional Neural Networks}, 
  year={2022},
  volume={15},
  number={},
  pages={6789-6803},
  doi={10.1109/JSTARS.2022.3198475}}

@ARTICLE{Liu2022,
  author={Liu, Guangyang and Liu, Bin and Zheng, Gang and Li, Xiaofeng},
  journal={IEEE Transactions on Geoscience and Remote Sensing}, 
  title={Environment Monitoring of Shanghai Nanhui Intertidal Zone With Dual-Polarimetric SAR Data Based on Deep Learning}, 
  year={2022},
  volume={60},
  number={},
  pages={1-18},
  doi={10.1109/TGRS.2022.3197149}}

@inproceedings{Simonyan2015,
  title={Very Deep Convolutional Networks for Large-Scale Image Recognition},
  author={Simonyan, Karen and Zisserman, Andrew},
  booktitle={ICLR},
  year={2015}
}

@inproceedings{He2016,
  title={Deep Residual Learning for Image Recognition},
  author={He, Kaiming and Zhang, Xiangyu and Ren, Shaoqing and Sun, Jian},
  booktitle={CVPR},
  pages={770--778},
  year={2016}
}

\appendix
\balance
\section{Open science}

\label{S_data_availability}

As a supplementary artifact, we provide the whole code and sufficient data to fully reproduce the findings of this article:

\begin{itemize}

    \item \textbf{Code.} The code used for this article is available to reviewers at the following link: \url{https://anonymous.4open.science/r/QuantumSAR}.

    \item \textbf{Filtered IQ dataset.} The full set of filtered IQ recordings collected from the 37 operational ICEYE satellites (post-filtering, as described in Section \ref{ss_pre_processing}). These files are sufficient to reproduce all spectrograms, training runs, and reported F1-scores. The dataset is hosted anonymously and linked from the anonymous repository above. Available at: \url{https://drive.proton.me/urls/470HH26X9G#MCHdOH2JkSXS}

\end{itemize}

Both the repository and the dataset will be transferred to Zenodo upon acceptance.

\section*{AI use disclosure}

Generative AI tools, including ChatGPT and Claude, were used to assist with manuscript writing and linguistic refinement, as well as with code development, debugging, and documentation. All AI-assisted content and code were reviewed and validated by the authors. The scientific methodology, experimental design, analysis, interpretation of results, and conclusions remain the responsibility of the authors, who take full responsibility for the final manuscript and associated code.

\end{document}